\makeatletter
\disable@package@load{breakurl}{}
\makeatother
\documentclass[sn-mathphys-num]{sn-jnl}
\usepackage{graphicx}%
\usepackage[T1]{fontenc}
\usepackage{booktabs}
\usepackage[table,xcdraw]{xcolor}
\usepackage{multirow}%
\usepackage{amsmath,amssymb,amsfonts}%
\usepackage{amsthm}%
\usepackage{mathrsfs}%
\usepackage[title]{appendix}%
\usepackage{xcolor}%
\usepackage{textcomp}%
\usepackage{manyfoot}%
\usepackage{algorithm}%
\usepackage{algorithmicx}%
\usepackage{algpseudocode}%
\usepackage{listings}%
\usepackage{bm}
\usepackage{pifont}
\usepackage{makecell}
\usepackage{array}
\usepackage{float}
\usepackage{multirow}
\usepackage[normalem]{ulem}
\usepackage{color, colortbl}
\usepackage{tabstackengine}
\usepackage{flushend}
\usepackage{balance}
\usepackage{threeparttable}
\usepackage{tabularx}

\usepackage{hyperref}

\theoremstyle{thmstyleone}%
\theoremstyle{thmstyletwo}%

\theoremstyle{thmstylethree}%

\begin{document}
\definecolor{myco}{RGB}{236,234,223}
\definecolor{myco2}{RGB}{216,214,194}

\title[Article Title]{Inductive-Associative Meta-learning Pipeline with Human Cognitive Patterns for Unseen Drug-Target Interaction Prediction}

\author[1]{\fnm{Xiaoqing} \sur{Lian}}
\author[1]{\fnm{Tianxu} \sur{Lv}}
\author[1]{\fnm{Jie} \sur{Zhu}}

\author[1]{\fnm{Shiyun} \sur{Nie}}

\author[1]{\fnm{Hang} \sur{Fan}}

\author[2]{\fnm{Guosheng} \sur{Wu}}

\author[2]{\fnm{Yunjun} \sur{Ge}}

\author*[3]{\fnm{Lihua} \sur{Li}}\email{lilh@hdu.edu.cn}

\author*[4]{\fnm{Xiangxiang} \sur{Zeng}}\email{xzeng@hun.edu.cn}

\author*[1,5]{\fnm{Xiang} \sur{Pan}}\email{xiangpan@jiangnan.edu.cn}

\affil[1]{\orgdiv{School of Artificial Intelligence and Computer Science}, \orgname{Jiangnan University}, \orgaddress{\city{Wuxi} \postcode{214122}, \state{Jiangsu}, \country{China}}}

\affil[2]{\orgdiv{Department of Basic Medical Science,Wuxi School of Medicine}, \orgname{Jiangnan University}, \orgaddress{\city{Wuxi} \postcode{214122}, \state{Jiangsu}, \country{China}}}

\affil[3]{\orgdiv{Institute of Biomedical Engineering and Instrumentation}, \orgname{Hangzhou Dianzi University}, \orgaddress{\city{Hangzhou} \postcode{310018}, \state{Zhejiang}, \country{China}}}

\affil[4]{\orgdiv{College of Information Science and Engineering}, \orgname{Hunan University}, \orgaddress{\city{Changsha}, \postcode{410082} \state{Hunan}, \country{China}}}

\affil[5]{\orgdiv{Shanghai Key Laboratory of Molecular Imaging}, \orgname{Shanghai University of Medicine and Health Sciences}, \orgaddress{\city{Shanghai} \postcode{201318}, \country{China}}}


\abstract{Significant differences in protein structures hinder the generalization of existing drug-target interaction (DTI) models, which often rely heavily on pre-learned binding principles or detailed annotations. In contrast, BioBridge designs an Inductive-Associative pipeline inspired by the workflow of scientists who base their accumulated expertise on drawing insights into novel drug-target pairs from weakly related references. BioBridge predicts novel drug-target interactions using limited sequence data, incorporating multi-level encoders with adversarial training to accumulate transferable binding principles. On these principles basis, BioBridge employs a dynamic prototype meta-learning framework to associate insights from weakly related annotations, enabling robust predictions for previously unseen drug-target pairs. Extensive experiments demonstrate that BioBridge surpasses existing models, especially for unseen proteins. Notably, when only homologous protein binding data is available, BioBridge proves effective for virtual screening of the epidermal growth factor receptor and adenosine receptor, underscoring its potential in drug discovery.
}

\maketitle
\section{Introduction}\label{sec1}
Identifying drug-target interactions (DTIs) is a cornerstone of drug discovery and development, significantly influencing various biological processes~\cite{luo2017network}. Traditional methods rely heavily on labor-intensive and fund-intensive experimental analyses of chemical compounds~\cite{broach1996high,bakheet2009properties}, which are increasingly complemented by computational approaches~\cite{chen2020transformercpi,wang2021quantitative,faulon2008genome}. Current computational methods include regression models predicting interaction strengths (e.g., Ki, IC50) and classification models identifying binding interactions based on potency thresholds~\cite{chatterjee2023improving,chen2020transformercpi}. 

DTI prediction methods typically fall into two categories: inductive and analogical. Inductive models utilize representations such as 3D structures or 2D sequences~\cite{bagherian2021machine} and employ advanced architectures like graph neural networks (GNNs) to generate embeddings for drug-protein pairs~\cite{koh2023psichic,lu2022tankbind,li2022heterogeneous,huang2021moltrans,nguyen2021graphdta,tsubaki2019compound}. Although effective under controlled conditions, these models often fail to generalize to novel drug-target pairs, as they primarily memorize existing annotations rather than identifying universal interaction mechanisms~\cite{chen2019hidden,gao2018interpretable}. Analogical methods focus on discovering correlations using fine-grained interaction annotations related to the drug-target pairs of interest~\cite{tian2022meta,wang2023zerobind}. Despite their promise, these methods require highly specific data, making new drug development difficult. A key limitation in both approaches is the significant variation among proteins, which hampers the transferability of learned drug-target binding mechanisms to novel interactions.

The drug development process by scientists employs a combination of inductive reasoning and associative strategies. Inductive reasoning involves understanding and applying established criteria for drug-target interactions, while associative strategies include consulting relevant literature to inform the process. This dual approach underscores the multifaceted nature of drug discovery, which extends beyond the mere application of existing knowledge or reliance on external data sources~\cite{macossay2019balancing,gurusinghe2022cold}.

Drawing inspiration from this workflow, we propose BioBridge, an inductive-associative pipeline designed to predict novel drug-target interactions with high accuracy and minimal cost. BioBridge operates in two stages: induction and association. In the induction phase, a multi-scale perception encoder identifies binding patterns across different levels, while adversarial learning strategies filter transferable binding principles. In the association phase, BioBridge mimics the literature review process by employing clustering-based task partitioning and designs a dynamic prototypical meta-learning algorithm to infer reliable interactions from limited binding annotations.

In a proof-of-concept study, we conducted extensive experiments on cold-pair, cross-domain zero-shot, and few-shot split, showing that BioBridge consistently outperforms other methods. Particularly for novel drug-target pairs, BioBridge demonstrates a significant enhancement of up to 30$\%$ over inductive methods by including transferable interaction mechanisms and extracting insights from sparse, weakly correlated interaction annotations. Interaction predictions of epidermal growth factor and adenosine receptor family further validate BioBridge's reliability with limited consistent protein-drug data. Ablation studies and interpretability analyses also offer valuable insights for future research.

\begin{figure*}[!ht]
    \centering
    \includegraphics[width=0.95\textwidth]{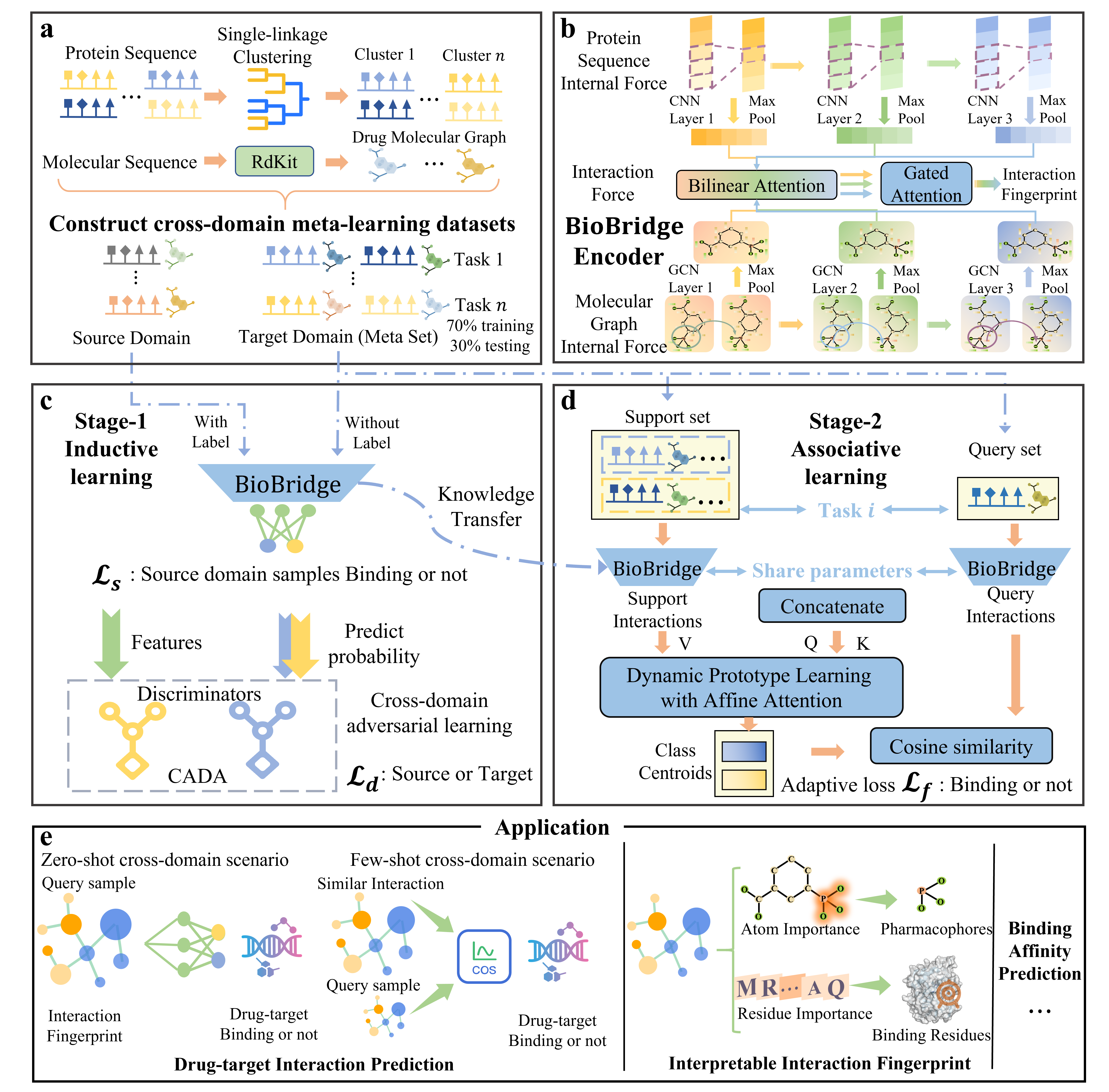}
    \caption{(a) Dataset Preparation: Protein sequences are clustered into $n$ classes. Drug molecular sequences are represented as molecular graphs via the RdKit~\cite{landrum2013rdkit}. Based on protein clusters, data is split into source and target domains, then divided into meta-tasks. (b) BioBridge Encoder: BioBridge inputs protein sequences and molecular graphs using CNN and GCN to model internal forces. Bilinear attention captures interactions at multiple levels, while gated attention aggregates interpretable interaction fingerprints. (c) Stage One: BioBridge pre-trains with labelled source data and unlabeled target data. The loss function $\mathcal{L}_s$ learns binding information, while $\mathcal{L}_d$ handles cross-domain adversarial learning. This binding knowledge is transferred to the next stage. (d) Stage Two: Tasks from the target domain, form support and query sets. The concatenated interactions are treated as Q and K, and support interactions as V. A dynamic prototype learning module defines unique class prototypes. Cosine similarity determines binding status, with an adaptive loss function $\mathcal{L}_f$ facilitating learning. (e) BioBridge generalizes well across tasks, providing interpretable interaction fingerprints for biological insights.}\label{fig:1}
\end{figure*}

\section{Results}\label{sec2}
\subsection{BioBridge pipeline}
BioBridge predicts DTIs by leveraging protein sequences and molecular graphs derived from drug sequences (Fig~\ref{fig:1}(a)). It trains on annotated data, including binding affinities and functional effects, framing interaction occurrence predictions as binary classifications. For validation, proteins are clustered into known (source) and unknown (target) domains, with the target domain split into training and testing sets based on clustering in the meta-learning setup. The cross-domain configuration emulates the data distribution encountered in real-world drug development. Meta-tasks delineated by clustering approximate the process scientists undergo when reviewing pertinent literature. Section~\ref{secA5} details our approach to removing data redundancy. 

BioBridge highlights the importance of multi-level understanding for drug target prediction and notes that shallow model features are more readily transferable to novel domains, both of which motivates the use of multi-level protein and molecular features to identify universal binding patterns (Fig~\ref{fig:1}(b)). BioBridge interweaves intra-molecular and intra-protein forces through three convolutional layers, employing max-pooling to extract fundamental atoms and residues. BioBridge simulates ligand-protein interactions through Bilinear Attention at every level, using a gated mechanism to consolidate interactions into a unified representation. 

BioBridge employs cross-domain adversarial learning to synchronize positive and negative interaction information between known and unknown pairs, capturing universally applicable interaction mechanisms for unknown drug-target pairs during inductive-learning stage(Fig~\ref{fig:1}(c)). These identified patterns are imparted to the meta-learning model by transferring model parameters. To avoid confusion between positive and negative interaction patterns in the unknown domain, BioBridge differentiates between these interactions during the adversarial learning phase. 

BioBridge designs a meta-learning approach to forecast novel drug-target interactions by leveraging pre-learned transferable binding principles and drawing inspiration from associative annotations of non-consistently paired interactions(Fig~\ref{fig:1}(d)). BioBridge uses a siamese network to create interaction prototypes for positive and negative interactions, categorizing new queries by cosine similarity. Predictions are refined using an affine attention-based dynamic prototype algorithm and an adaptive loss function. 

As depicted in Fig~\ref{fig:1}(e), the application of BioBridge encompasses the prediction of interactions for unknown drug-target pairs in zero-shot and few-shot learning scenarios. Furthermore, BioBridge provides biological interpretability through attention-based visualization. Beyond these capabilities, BioBridge is also adept at predicting drug-target binding affinities about the interaction between drugs and their targets.

To clarify the different models denoted by various notations during the training phases of BioBridge, refer to Table~\ref{tab:variantions} for an elucidation. Details of the experimental setup are provided in the Section~\ref{secA3}.

\begin{figure*}[h]
    \centering
    \includegraphics[width=\textwidth]{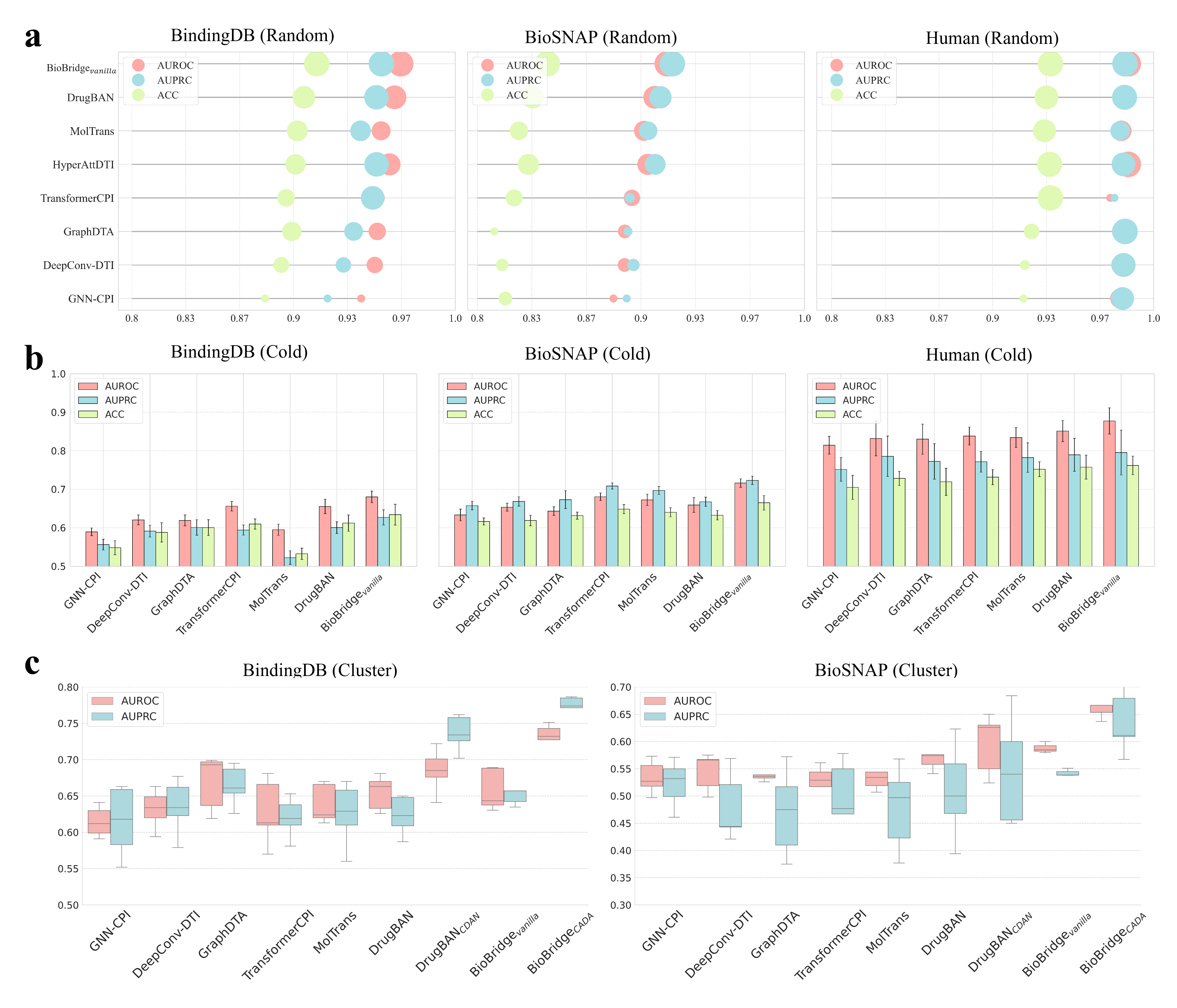}
    \caption{(a) Performance comparison on the BindingDB, BioSNAP, and Human datasets using random splitting.  The bubble size corresponds to the value of the metric. (b) Comparison of cross-domain performance on the BindingDB and BioSNAP datasets using cluster-based pair splitting. (c) Zero-shot comparison of meta cross-domain splitting on BindingDB and BioSNAP datasets, indicating that protein differences are important factors limiting drug target prediction.}\label{fig:2}
\end{figure*}

\begin{table*}[!ht]
    \renewcommand{\arraystretch}{1.2}
    \centering
    \rowcolors*{1}{myco}{white}
    \caption{Few shot comparison of meta unseen protein splitting on the BindingDB and BinSNAP datasets (\textbf{Best},\uline{ Second Best}).}\label{tab:2}
    \resizebox{\linewidth}{!}{
        \begin{tabular}{ccccccc}
            \Xhline{1.2pt} 
            Dataset & \multicolumn{3}{c}{BindingDB(Meta Unseen Protein)}                                      & \multicolumn{3}{c}{BioSNAP(Meta Unseen Protein)}                                        \\
            Metric                                  & AUROC                       & AUPRC                       & ACC                         & AUROC                       & AUPRC                       & ACC                         \\
            \Xhline{1.2pt}
            DrugBAN~\cite{bai2023interpretable} & $0.5573 \pm 0.0137^*$ & $0.5723 \pm 0.0181^*$ & $0.5281 \pm 0.0060^*$ & $0.5779 \pm 0.0175^*$& $0.7219 \pm 0.0203^*$& $0.4876 \pm 0.0125^*$ \\
            \rowcolor{myco2} Setting                & \multicolumn{6}{c}{1-shot}                                                                                                                                                        \\
            MAML++~\cite{MAMLPP}                    & $0.6200 \pm 0.0095$         & $0.6060 \pm 0.0053^*$         & $0.5894 \pm 0.0119^*$         & $0.5884 \pm 0.0136^*$         & $0.5659 \pm 0.0132^*$         & $0.5594 \pm 0.0086^*$         \\
            Protypes~\cite{snell2017prototypical}   & \uline{$0.7079\pm 0.0213^*$}  & \uline{$0.7080 \pm 0.0239$} & $0.6443\pm 0.0148$          & \uline{$0.6536\pm0.0114^*$}   & \uline{$0.6455\pm0.0136^*$}   & \uline{$0.6080\pm0.0071^*$}   \\
            MetaOptNet~\cite{lee2019meta}           & $0.7061 \pm 0.0033^*$         & $0.6898 \pm 0.0025^*$         & \uline{$0.6481 \pm 0.0103^*$} & $0.6429 \pm 0.0225$         & $0.6254 \pm 0.0225^*$         & $0.6053 \pm 0.0161$         \\
            ANIL~\cite{raghu2019rapid}              & $0.7046 \pm 0.0073$         & $ 0.6994 \pm 0.0082$        & $0.6441 \pm 0.0034^*$         & $0.6476 \pm 0.0067^*$         & $0.6379 \pm 0.0095^*$         & $0.6060 \pm 0.0054^*$         \\
            BioBridge$_{Meta}$                          & \bm{$0.7488\pm 0.0073$}     & \bm{$0.7545\pm 0.0075$}     & \bm{$0.6631\pm 0.0051$}     & \bm{$0.6755\pm 0.0073$}     & \bm{$0.6707\pm 0.0073$}     & \bm{$0.6237\pm 0.0040$}     \\
            \rowcolor{myco2} Setting                & \multicolumn{6}{c}{3-shot}                                                                                                                                                        \\
            MAML++~\cite{MAMLPP}                    & $0.7258 \pm 0.0194$         & $0.7146 \pm 0.0208$         & $0.6670 \pm 0.0156^*$         & $0.685 \pm 0.038^*$           & $0.6689 \pm 0.0424^*$         & $0.6362 \pm 0.0255^*$         \\
            Protypes~\cite{snell2017prototypical}   & $0.7838\pm 0.0149^*$          & $0.7840\pm 0.0172^*$          & $0.7028\pm 0.0128^*$          & $0.7352\pm0.0055^*$           & $0.7295\pm 0.0036^*$          & $0.6694\pm 0.0088^*$          \\
            MetaOptNet~\cite{lee2019meta}           & \uline{$0.8099 \pm 0.0136^*$} & \uline{$0.7903 \pm 0.0160^*$} & \bm{$0.7392 \pm 0.0108^*$} & \uline{$0.7603 \pm 0.0087^*$} & \uline{$0.7422 \pm 0.0058^*$} & \bm{$0.6975 \pm 0.0123^*$}    \\
            ANIL~\cite{raghu2019rapid}              & $0.7704 \pm 0.0182^*$         & $0.7682 \pm 0.0098$         & $0.6964 \pm 0.0107^*$         & $0.7116 \pm 0.0031^*$         & $0.695 \pm 0.0016^*$          & $0.6612 \pm 0.0024^*$         \\
            BioBridge$_{Meta}$                          & \bm{$0.8227\pm 0.0023$}     & \bm{$0.8227\pm 0.0029$}     & \uline{$0.7314\pm 0.0041$}  & \bm{$0.7661\pm 0.0076$}     & \bm{$0.7664\pm 0.0084$}     & \uline{$0.6910\pm0.0057$}   \\
            \rowcolor{myco2} Setting                & \multicolumn{6}{c}{5-shot}                                                                                                                                                        \\
            MAML++~\cite{MAMLPP}                    & $0.7452 \pm 0.0196^*$         & $0.7392 \pm 0.0243^*$         & $0.6801 \pm 0.0135^*$         & $0.7251 \pm 0.0341^*$         & $0.7055 \pm 0.0353^*$         & $0.6682 \pm 0.0244^*$         \\
            Protypes~\cite{snell2017prototypical}   & $0.8200\pm 0.0104^*$          & $0.8205\pm 0.0114^*$          & $0.7353\pm 0.0085^*$          & $0.7718\pm 0.0058^*$          & $0.7674\pm 0.0073^*$          & $0.6983\pm 0.0031^*$          \\
            MetaOptNet~\cite{lee2019meta}           & \uline{$0.8524 \pm 0.0068$} & \uline{$0.8381 \pm 0.0082^*$} & \bm{$0.7703 \pm 0.0073$}    & \uline{$0.7880 \pm 0.0050^*$} & \uline{$0.7734 \pm 0.0027^*$} & \uline{$0.7111 \pm 0.0059^*$} \\
            ANIL~\cite{raghu2019rapid}              & $0.7933 \pm 0.0019^*$         & $0.7784 \pm 0.0089^*$         & $0.7209 \pm 0.0122^*$         & $ 0.7635 \pm 0.013^*$         & $0.7560 \pm 0.0143^*$         & $0.6934 \pm 0.0121$         \\
            BioBridge$_{Meta}$                          & \bm{$0.8527\pm 0.0006$}     & \bm{$0.8555\pm 0.0007$}     & \uline{$0.7624\pm 0.0016$}  & \bm{$0.8030\pm 0.0006$}     & \bm{$0.8052\pm 0.0022$}     & \bm{$0.7235\pm 0.0045$} \\
            \Xhline{1.2pt}   
            \rowcolor{white} \multicolumn{7}{l}{\small $*$ Significantly different (p < 0.05) from the corresponding BioBridge metric value; one-way analysis of variance (ANOVA).}\\
            \end{tabular}
    }
\end{table*}

\subsection{BioBridge induces transferable interaction principles}
Zero-shot prediction of drug-target interactions is a pivotal evaluation task in modern drug discovery, as it measures a model's ability to capture established drug-target binding principles and generalize to unseen scenarios.   To assess the inductive capabilities of BioBridge, we conducted experiments on three prominent datasets: BindingDB, BioSNAP, and Human.   Recognizing the importance of addressing real-world challenges in novel drug development, we adopted both cold-pair splits (Cold) and cross-domain splits (Cluster) to simulate unknown scenarios.   For the cross-domain splits, BioBridge utilized the Category-Aware Domain Adversarial Learning (CADA) module to enhance its predictive accuracy.

As shown in Fig~\ref{fig:2}(a)(b) and Table~\ref{tab:a4}, BioBridge consistently outperformed state-of-the-art models that prioritize deep feature extraction, particularly in predicting interactions for unseen pairs. However, BioBridge's performance on random splits was less pronounced, indicating that while deep features effectively model known drug-target interactions, shallow features demonstrate superior inductive capabilities for addressing unknown interaction mechanisms.

In cross-domain evaluations, we incorporated the CADA module, forming BioBridge$_{CADA}$. For fair comparison, DrugBAN$_{CDAN}$\cite{bai2023interpretable,long2018conditional}, optimized for cross-domain tasks, was also included. Due to the limited size of the Human dataset, experiments focused on the BindingDB and BioSNAP datasets, employing cluster-based splitting\cite{bai2023interpretable} to simulate cross-domain scenarios. As illustrated in Fig~\ref{fig:2}(c), performance declined across all models when evaluated on entirely novel drug-target pairs. Despite this challenge, BioBridge demonstrated robust performance, surpassing non-cross-domain models on both datasets. BioBridge$_{CADA}$ achieved the highest results by effectively distinguishing domain differences and capturing transferable interaction patterns. This highlights the strength of combining a multi-level encoder with the CADA module in concluding transferable drug-target interaction principles.

To further validate our approach, we compared BioBridge against structure-based protein-ligand models on the PDB2020 dataset\cite{lu2022tankbind,stark2022equibind}. Table~\ref{tab:a5} shows that BioBridge achieves leading results among sequence-based model and is comparable to structure-based models, underscoring its adaptability and effectiveness even in data-limited settings. 

\subsection{BioBridge derives inspiration by association from weakly correlated annotations}
BioBridge inherits the transferable interaction principles learned during the inductive stage. To emulate the scientific process of drawing insights from references, we use meta-learning to activate BioBridge's associative potential. Our first objective is to quantitatively identify the factors contributing to prediction errors for novel drug-target pairs, which will guide the focus of subsequent experiments.  To this end, we conducted evaluations under two distinct scenarios: cross-domain novel drugs (Meta Unseen Drug) and cross-domain novel proteins (Meta Unseen Protein). Figure~\ref{fig:2}(c) illustrates that BioBridge$_{CADA}$ outperforms other models in generalization. Predicting interactions for entirely unknown proteins proved more challenging than for novel molecules, with performance differing by nearly 20$\%$. This highlights a critical challenge in drug-target prediction: the variability among proteins. Consequently, our few-shot experiments focus on unseen proteins, employing BioBridge$_{CADA}$'s final epoch parameters to ensure effective knowledge transfer.

Few-shot learning assesses a model's ability to generalize when each class contains only $n$ samples. Due to the limited size of the Human dataset, we focused on associative learning using the BindingDB and BioSNAP datasets. To derive insights from weakly related reference annotations for the target drug-target pairs, we designed few-shot tasks based on clustering target categories rather than individual classes (Section \ref{sec4_3}). By integrating the dynamic prototype learning module with the BioBridge encoder, we developed BioBridge$_{Meta}$. For comparison, we also trained DrugBAN on both source and target training domain datasets and evaluated it on the target test domain.

Across both datasets, BioBridge$_{Meta}$ achieved optimal or near-optimal results on all metrics, with AUROC improvements of up to 30$\%$ in BindingDB, significantly outperforming inductive baseline methods (Table~\ref{tab:2}). Notably, DrugBAN performed worse when trained on the target domain training set than when trained solely on the source domain, illustrating the shortcut learning tendency of inductive methods, which often memorize annotations instead of learning intrinsic interactions~\cite{bai2023interpretable}. The superior performance of BioBridge$_{Meta}$ stems from its dynamic prototype algorithm, which effectively addresses protein differences and extracts meaningful information from weakly related reference annotations. This adaptability also explains the strong performance of MetaOptNet, which incorporates mechanisms for adaptive prototype construction. Additionally, in tasks where the support and query sets involved the same protein, BioBridge demonstrates remarkable performance, as shown in Table~\ref{tab:3}. This setup simplifies task complexity but demands greater data collection effort, making BioBridge's success particularly notable.

In summary, the BioBridge pipeline builds upon generalized binding principles established during the inductive phase and leverages weakly associated annotations for associative learning, enabling accurate predictions of unknown drug-target interactions. Additionally, BioBridge relies exclusively on sequence data, allowing efficient and rapid training and deployment tailored to specific scenarios (Section \ref{secA6}). These features underscore its potential for practical applications in drug discovery.

\begin{figure*}[!htbp]
    \centering
    \includegraphics[width=0.9\linewidth]{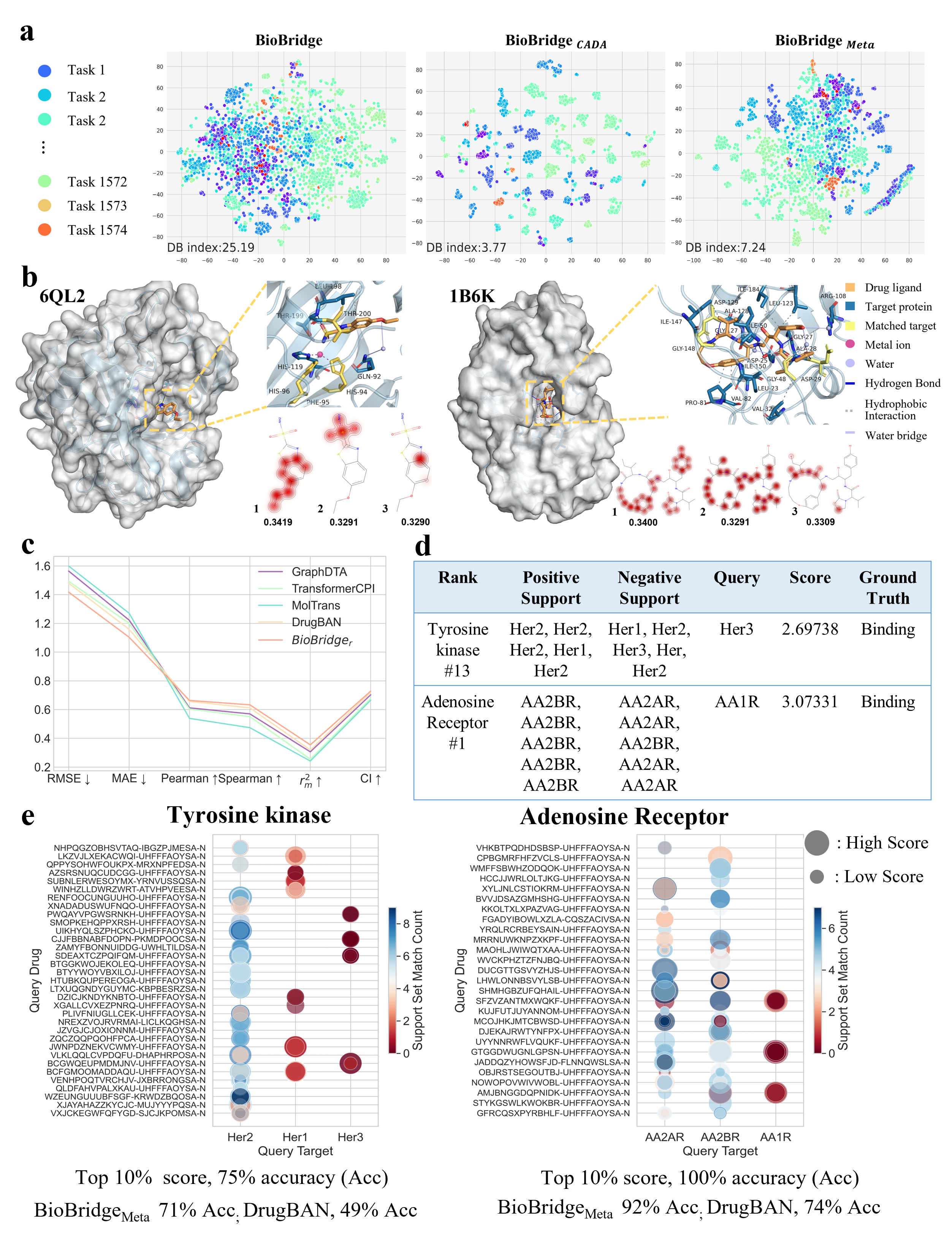}
    \caption{(a) The t-SNE visualization presents drug-target pairs across various tasks, with lower Davies-Bouldin (DB) indices indicating superior performance. (b) The visualization of drug ligand and target binding pocket attention is generated using rdkit~\cite{landrum2013rdkit} for drug molecule mapping and PyMOL~\cite{delano2002pymol} for target plotting. PLIP~\cite{adasme2021plip} is utilized to plot the forces between drug targets. The molecular diagram highlights the top 20$\%$ of model-concerned positions in red, while the target docking diagram depicts the model's focus in yellow, aligning with actual interacting residues. (c) Performance on the PDBBind v2020 datasets is compared, with $\uparrow$ indicating higher scores are favorable and $\downarrow$ signifying the opposite. (d) Virtual screening examples from tyrosinase and adenosine receptor families feature top 10$\%$ scoring compounds with a limited query protein in the support set. (e) In these virtual screenings, color represents the query protein's frequency in the support set, and bubble size corresponds to score magnitude. Top 10$\%$ meta-tasks for adenosine receptors even achieved 100$\%$ accuracy. BioBridge surpassed DrugBAN in traditional classification tasks, demonstrating superior adaptability to novel drug-target pairs.}\label{fig:3}
\end{figure*}

\subsection{Interpretability analysis}
How does BioBridge benefit from both inductive and associative learning? Has BioBridge captured the fundamental chemical principles that determine binding? To address these questions, we visualized the learned representations of BioBridge and conducted interpretative analyses of its results on two crystal structures that were unseen during training.

To evaluate the impact of cross-domain pre-training on knowledge transfer, we compare the performance of three vanilla BioBridge, BioBridge$_{CADA}$, and BioBridge$_{Meta}$—in a 5-shot scenario, using the BindingDB dataset to extract drug-target binding features for an unknown protein task. We apply the t-SNE algorithm~\cite{t_SNE} to reduce these features to two dimensions, as shown in Fig~\ref{fig:3}(a). The Davies-Bouldin index~\cite{petrovic2006comparison} is used to assess the classification effectiveness across meta-tasks. Our results show that vanilla BioBridge struggles to distinguish protein types based on drug-binding properties, highlighting the challenges of generalizing to new tasks. In contrast, BioBridge$_{CADA}$, which incorporates the CADA module, effectively identifies variations in binding modes, improving knowledge transfer to new protein targets. Although BioBridge$_{Meta}$ exhibits a slight decrease in protein discrimination after meta-training, it still outperforms vanilla BioBridge, likely due to its focus on associative ability during meta-training.

BioBridge uses attention mechanisms to encode drug-target interactions, enabling the model to analyze the impact of substructures on binding. We examined high-resolution X-ray structures of human protein targets from PDB entries 6QL2~\cite{kazokaite2019engineered} and 1B6K~\cite{martin1999molecular}, selecting structures with co-crystallized ligands (pIC50 $\leq$ 100 nM) that were not part of the training set. Fig~\ref{fig:3}(b) shows the ligand-protein interaction maps, with the top 20$\%$ of atoms, based on bilinear attention coefficients, highlighted in red across the model's three hierarchical domains. Residues of the target protein that align with the X-ray structures are marked in yellow.

For 6QL2, the first domain focuses on the benzene ring and ethoxy group, which engage in hydrophobic interactions and potential water-mediated bridging. The second domain highlights the sulfonyl region, emphasizing sulfur interactions. The third domain targets the benzothiazole bridgehead carbon atom, revealing no specific interactions with the protein. For 1B6K, BioBridge's first domain identifies the heterocyclic bi-ring of PI5, with hydrophobic interactions and hydrogen bonding. The second domain detects hydrogen bond donors and notes hydrophobic interactions, while the third domain involves the nitrogen atom as a hydrogen bond donor, with oxygen acceptors and water-mediated bridging interactions. Despite the challenges in interpreting protein sequences, BioBridge successfully identifies key residues involved in binding for both 6QL2 and 1B6K, demonstrating its potential for capturing interaction principles for unknown drug-target pairs.
\begin{table*}[!ht]
    \renewcommand{\arraystretch}{1.2}
    \centering
    \rowcolors*{2}{white}{myco}
    \caption{Ablation experiment of BioBridge encoder on meta unseen protein splitting of BindingDB and BioSNAP dataset(\textbf{Best},\uline{ Second Best}).}\label{tab:7}
    \resizebox{\linewidth}{!}{
    \begin{tabular}{ccccccccccccc}
        \Xhline{1.2pt}
        \rowcolor{myco2}
    \multicolumn{4}{c}{Dataset}                     & \multicolumn{9}{c}{BindingDB (Meta unseen protein)}                                                                                                                                                                                                 \\ 
    \multicolumn{4}{c}{Setting}                   & \multicolumn{3}{c}{1- shot}                                                    & \multicolumn{3}{c}{3- shot}                                                  & \multicolumn{3}{c}{5- shot}                                                   \\ 
    MA        & DGL       & FL        & Load      & \multicolumn{1}{c}{AUROC} & \multicolumn{1}{c}{AUPRC} & \multicolumn{1}{c}{ACC} & \multicolumn{1}{c}{AUROC} & \multicolumn{1}{c}{AUPRC} & \multicolumn{1}{c}{ACC} & \multicolumn{1}{c}{AUROC} & \multicolumn{1}{c}{AUPRC} & \multicolumn{1}{c}{ACC} \\ 
    \Xhline{1.2pt}
    \ding{51} & \ding{51} & \ding{51} & \ding{51} & \bm{$0.7488\pm 0.0073$}        & \bm{$0.7545\pm 0.0075$}        & $0.6631\pm 0.0051$      & \bm{$0.8227\pm 0.0023$}        & \bm{$0.8227\pm 0.0029$}        & \bm{$0.7314\pm 0.0041$}      & \bm{$0.8527\pm 0.0006$}           & \bm{$0.8555\pm 0.0007$}        & \bm{$0.7624\pm 0.0016$}      \\
    \ding{51} & \ding{55} & \ding{51} & \ding{51} & \uline{$0.7412\pm 0.0101$}        & \uline{$0.7467\pm 0.0104$}        & \uline{$0.6659\pm 0.0075$}      & \uline{$0.8205\pm 0.0034$}        & \uline{$0.8245\pm 0.0052$}        & \uline{$0.7303\pm 0.0065$}      & \uline{$0.8465\pm 0.0031$}        &\uline{ $0.8489\pm 0.0028$}       & $0.7535\pm 0.0042$      \\
    \ding{51} & \ding{51} & \ding{55} & \ding{51} & $0.7194\pm 0.0182$        & $0.7212\pm 0.0210$        & $0.6541\pm 0.0098$      & $0.7966\pm 0.0031$        & $0.7991\pm 0.0003$        & $0.7132\pm 0.0052$      & $0.8306\pm 0.0130$        & $0.8322\pm 0.0104$        & $0.7458\pm 0.0099$      \\
    \ding{51} & \ding{55} & \ding{55} & \ding{51} & $0.7079\pm 0.0213$        & $0.7080\pm 0.0239$        & $0.6443\pm 0.0148$      & $0.7838\pm 0.0149$        & $0.7840\pm 0.0172$        & $0.7028\pm 0.0128$      & $0.8200\pm 0.0104$        & $0.8205\pm 0.0114$        & $0.7353\pm 0.0085$      \\
    \ding{51} & \ding{51} & \ding{51} & \ding{55} & $0.7232\pm 0.0042$        & $0.7258\pm 0.0054$        & $0.6531\pm 0.0036$      & $0.7924\pm 0.0043$        & $0.7945\pm 0.0040$        & $0.7091\pm 0.0049$      & $0.8351\pm 0.0085$        & $0.8351\pm 0.0071$        & $0.7483\pm 0.0106$      \\
    \ding{51} & \ding{51} & \ding{55} & \ding{55} & $0.6777\pm 0.0083$        & $0.6743\pm 0.0091$        & $0.6224\pm 0.0076$      & $0.7647\pm 0.0234$        & $0.7652\pm 0.0225$        & $0.6880\pm 0.0183$      & $0.7822\pm 0.1417$        & $0.7808\pm 0.0176$        & $0.7055\pm 0.0095$      \\
    \ding{51} & \ding{55} & \ding{51} & \ding{55} & $0.7279\pm 0.0109$        & $0.7304\pm 0.0105$        & $0.6557\pm 0.0091$      & $0.7880\pm 0.0107$        & $0.7899\pm 0.0109$        & $0.7063\pm 0.0081$      & $0.8259\pm 0.0048$        & $0.8259\pm 0.0042$        & $0.7407\pm 0.0067$      \\
    \ding{51} & \ding{55} & \ding{55} & \ding{55} & $0.6983\pm 0.0074$        & $0.6996\pm 0.0085$        & $0.6354\pm 0.0065$      & $0.7498\pm 0.0190$        & $0.7492\pm 0.0175$        & $0.6783\pm 0.0157$      & $0.7760\pm 0.0165$        & $0.7726\pm 0.0201$        & $0.7022\pm 0.0093$      \\
    \ding{55} & \ding{51} & \ding{51} & \ding{51} & $0.7388 \pm  0.0062$      & $0.7463 \pm  0.0103$      & $0.6629 \pm  0.0019$    & $0.8140 \pm  0.0011$      & $0.8135 \pm  0.0055$      & $0.7292 \pm  0.0010$    & $0.8433 \pm  0.0002$      & $0.8448 \pm  0.0002$      & \uline{$0.7547 \pm  0.0002$}    \\
    \ding{55} & \ding{55} & \ding{51} & \ding{51} & $0.7398 \pm  0.0090$      & $0.7385 \pm  0.0108$      & \bm{$0.6701 \pm  0.0048$}    & $0.8043 \pm  0.0066$      & $0.8028 \pm  0.0076$      & $0.7222 \pm  0.0056$    & $0.8349 \pm  0.0042$      & $0.8362 \pm  0.0030$       & $0.7466 \pm  0.0043$    \\
    \ding{55} & \ding{51} & \ding{55} & \ding{51} & $0.7063 \pm  0.0256$      & $0.7027 \pm  0.0284$      & $0.6475 \pm  0.0162$    & $0.7979 \pm  0.0023$      & $0.7974 \pm  0.0004$      & $0.7162 \pm  0.0051$    & $0.8205 \pm  0.0075$      & $0.8204 \pm  0.0072$      & $0.7358 \pm  0.0067$    \\
    \ding{55} & \ding{55} & \ding{55} & \ding{51} & $0.7355 \pm  0.0049$      & $0.7362 \pm  0.0065$      & $0.6641 \pm  0.0032$    & $0.7878 \pm  0.0008$      & $0.7833 \pm  0.0001$      & $0.7108 \pm  0.0012$    & $0.8033 \pm  0.0061$      & $0.8002 \pm  0.0038$      & $0.7257 \pm  0.0066$    \\
    \ding{55} & \ding{51} & \ding{51} & \ding{55} & $0.7191 \pm  0.0172$      & $0.7201 \pm  0.0193$      & $0.6508 \pm  0.0118$    & $0.7988 \pm  0.0012$      & $0.8003 \pm  0.0030$       & $0.7154 \pm  0.0007$    & $0.8272 \pm  0.0039$      & $0.8266 \pm  0.0042$      & $0.7413 \pm  0.0018$    \\
    \ding{55} & \ding{51} & \ding{55} & \ding{55} & $0.6660  \pm  0.0062$      & $0.6590 \pm  0.0067$      & $0.6151 \pm  0.0044$    & $0.7362 \pm  0.0068$      & $0.7296 \pm  0.0074$      & $0.6702 \pm  0.0039$    & $0.7718 \pm  0.0088$      & $0.7671 \pm  0.0096$      & $0.6971 \pm  0.0073$    \\
    \ding{55} & \ding{55} & \ding{51} & \ding{55} & $0.7119 \pm  0.0087$      & $0.7115 \pm  0.0098$      & $0.6451 \pm  0.0074$    & $0.7867 \pm  0.0048$      & $0.7865 \pm  0.0042$      & $0.7049 \pm  0.0060$     & $0.8194 \pm  0.0027$      & $0.8193 \pm  0.0035$      & $0.7344 \pm  0.0023$    \\
    \ding{55} & \ding{55} & \ding{55} & \ding{55} & $0.6772 \pm  0.0029$      & $0.6704 \pm  0.0044$      & $0.6239 \pm  0.0021$    & $0.7053 \pm  0.0023$      & $0.6993 \pm  0.0007$      & $0.6465 \pm  0.0043$    & $0.7505 \pm  0.0028$      & $0.7447 \pm  0.0006$      & $0.6828 \pm  0.0020$     \\ \Xhline{1.2pt}

    \rowcolor{myco2}
    \multicolumn{4}{c}{Dataset}                     & \multicolumn{9}{c}{BioSNAP (Meta unseen protein)}                                                                                                                                                                                                                 \\ 
    \multicolumn{4}{c}{Setting}                   & \multicolumn{3}{c}{1- shot}                                                        & \multicolumn{3}{c}{3- shot}                                                       & \multicolumn{3}{c}{5-shot}                                                          \\ 
    MA        & DGL       & FL        & Load      & \multicolumn{1}{c}{AUROC} & \multicolumn{1}{c}{AUPRC}  & \multicolumn{1}{c}{ACC}    & \multicolumn{1}{c}{AUROC}  & \multicolumn{1}{c}{AUPRC}  & \multicolumn{1}{c}{ACC}    & \multicolumn{1}{c}{AUROC}  & \multicolumn{1}{c}{AUPRC}  & \multicolumn{1}{c}{ACC}      \\
    \Xhline{1.2pt}
    \ding{51} & \ding{51} & \ding{51} & \ding{51} & \bm{$0.6755\pm 0.0073$}   & \uline{$0.6707\pm 0.0073$} & \uline{$0.6237\pm 0.0040$} & \bm{$0.7661\pm 0.0076$}    & \bm{$0.7664\pm 0.0084$}    & \bm{$0.6910\pm 0.0057$}    & \bm{$0.8030\pm 0.0006$}    & \bm{$0.8052\pm 0.0022$}    & \bm{$0.7215\pm 0.0045$}      \\
    \ding{51} & \ding{55} & \ding{51} & \ding{51} & \uline{$0.6755\pm 0.0076$}   & \bm{$0.6710\pm 0.0104$}    & \bm{$0.6238\pm 0.0065$}    & \uline{$0.7630\pm 0.0071$} & \uline{$0.7623\pm 0.0072$} & \uline{$0.6889\pm 0.0065$} & \uline{$0.7948\pm 0.0051$} & \uline{$0.7937\pm 0.0070$} & $0.7150\pm 0.0065$           \\
    \ding{51} & \ding{51} & \ding{55} & \ding{51} & $0.6542\pm 0.0111$        & $0.6456\pm 0.0133$         & $0.6042\pm 0.0057$         & $0.7435\pm 0.0046$         & $0.7397\pm 0.0019$         & $0.6779\pm 0.0048$         & $0.7781\pm 0.0060$         & $0.7746\pm 0.0069$         & $0.7045\pm 0.0015$           \\
    \ding{51} & \ding{55} & \ding{55} & \ding{51} & $0.6536\pm 0.0114$        & $0.6455\pm 0.0136$         & $0.6080\pm 0.0071$         & $0.7352\pm 0.0055$         & $0.7295\pm 0.0036$         & $0.6694\pm 0.0088$         & $0.7718\pm 0.0058$         & $0.7674\pm 0.0073$         & $0.6983\pm 0.0031$           \\
    \ding{51} & \ding{51} & \ding{51} & \ding{55} & $0.6731\pm 0.0097$        & $0.6675\pm 0.0110$         & $0.6252\pm 0.0043$         & $0.7434\pm 0.0165$         & $0.7400\pm 0.0178$         & $0.6753\pm 0.0113$         & $0.7766\pm 0.005$          & $0.7744\pm 0.0046$         & $0.7028\pm 0.0068$           \\
    \ding{51} & \ding{51} & \ding{55} & \ding{55} & $0.6532\pm 0.0236$        & $0.6468\pm 0.0243$         & $0.6072\pm 0.0173$         & $0.7024\pm 0.0162$         & $0.6977\pm 0.0175$          & $0.6463\pm 0.0123$         & $0.7246\pm 0.0409$         & $0.7206\pm 0.0428$          & $0.6620\pm 0.0296$           \\
    \ding{51} & \ding{55} & \ding{51} & \ding{55} & $0.6726\pm 0.0032$        & $0.6690\pm 0.0018$         & $0.6170\pm 0.0037$         & $0.7390\pm 0.0206$         & $0.7363\pm 0.0210$         & $0.6716\pm 0.0112$         & $0.7759\pm 0.0062$         & $0.7719\pm 0.0062$         & $0.7001\pm 0.0093$           \\
    \ding{51} & \ding{55} & \ding{55} & \ding{55} & $0.6496\pm 0.0165$        & $0.6432\pm 0.0152$         & $0.6026\pm 0.0116$         & $0.7029\pm 0.0319$         & $0.6986\pm 0.0311$         & $0.6444\pm 0.0224$         & $0.7329\pm 0.0242$         & $0.7309\pm 0.0259$         & $0.6649\pm 0.0197$           \\
    \ding{55} & \ding{51} & \ding{51} & \ding{51} & $0.6762 \pm  0.0133$      & $0.6691 \pm  0.0156$       & $0.6264 \pm  0.0078$       & $0.7585 \pm  0.0050$        & $ 0.7489 \pm  0.0035$      & $0.6933 \pm  0.0038$       & $0.7898 \pm  0.0015$       & $0.7803 \pm  0.0006$       & \uline{$0.7202 \pm  0.0006$} \\
    \ding{55} & \ding{55} & \ding{51} & \ding{51} & $0.6730 \pm  0.0219$      & $0.6645 \pm  0.0253$       & $0.6225 \pm  0.0160$        & $0.7492 \pm  0.0045$       & $ 0.7411 \pm  0.0065$      & $0.6838 \pm  0.0072$       & $0.7842 \pm  0.0033$       & $0.7756 \pm  0.0027$       & $0.7151 \pm  0.0025$         \\
    \ding{55} & \ding{51} & \ding{55} & \ding{51} & $0.6717 \pm  0.0189$      & $0.6608 \pm  0.0205$       & $0.6236 \pm  0.0152$       & $0.7386 \pm  0.0052$       & $ 0.7286 \pm  0.0058$      & $0.6769 \pm  0.0044$       & $0.7711 \pm  0.004$        & $0.7615 \pm  0.0074$       & $0.7078 \pm  0.0026$         \\
    \ding{55} & \ding{55} & \ding{55} & \ding{51} & $0.6772 \pm  0.0175$      & $0.6679 \pm  0.0202$       & $0.6249 \pm  0.0121$       & $0.7350 \pm  0.0063$       & $0.7241 \pm  0.0066$       & $0.6738 \pm  0.0057$       & $0.7613 \pm  0.004$        & $ 0.7498 \pm  0.0022$      & $0.6958 \pm  0.0022$         \\
    \ding{55} & \ding{51} & \ding{51} & \ding{55} & $0.6529 \pm  0.0079$      & $ 0.6454 \pm  0.0072$      & $0.6090 \pm  0.0098$       & $0.7353 \pm  0.0050$        & $0.7284 \pm  0.0062$       & $0.6689 \pm  0.0037$       & $0.7475 \pm  0.0067$       & $0.7391 \pm  0.0062$       & $0.6861 \pm  0.0017$         \\
    \ding{55} & \ding{51} & \ding{55} & \ding{55} & $0.6335 \pm  0.0258$      & $0.6280 \pm  0.0226$       & $0.5939 \pm  0.0276$       & $0.6835 \pm  0.0333$       & $0.6752 \pm  0.0276$       & $ 0.6296 \pm  0.0264$      & $0.7134 \pm  0.0201$       & $0.7065 \pm  0.0193$       & $0.6517 \pm  0.0156$         \\
    \ding{55} & \ding{55} & \ding{51} & \ding{55} & $0.6459 \pm  0.0078$      & $0.6395 \pm  0.0091$       & $0.6041 \pm  0.0030$       & $0.7260 \pm  0.0227$       & $0.7185 \pm  0.0258$       & $0.6628 \pm  0.0163$       & $0.7511 \pm  0.0108$       & $0.7425 \pm  0.0126$       & $0.6859 \pm  0.0036$         \\
    \ding{55} & \ding{55} & \ding{55} & \ding{55} & $0.6270 \pm  0.0217$      & $0.6188 \pm  0.0221$       & $0.5872 \pm  0.0175$       & $0.6709 \pm  0.0372$       & $0.6665 \pm  0.0369$       & $0.6184 \pm  0.0299$       & $0.7184 \pm  0.0080$        & $0.7130 \pm  0.0081$       & $0.6570 \pm  0.0044$         \\ \Xhline{1.2pt}
    \end{tabular}
    }
\end{table*}

\subsection{Virtual screening with BioBridge}
Having established the effectiveness and interpretability of BioBridge, we evaluate its practical application in virtual screening across two protein target families: the epidermal growth factor receptor family (HER1, HER2, HER3), crucial for cellular processes, and the adenosine receptor family (AA1R, AA2AR, AA2BR), essential for cellular signalling and immune responses. This evaluation demonstrates BioBridge's high extensibility in real-world scenarios, requiring chemists to reference only a few homologous protein binding interactions to predict novel drug-target pair bindings.

To enhance screening accuracy, we develop BioBridge$_r$, trained on PDB2020 data, integrating seamlessly with BioBridge via a single-layer decoding process. We also compare BioBridge$_r$ with several state-of-the-art models in PDB2020, as depicted in Fig~\ref{fig:3}(c). BioBridge$_r$ demonstrates comprehensive superiority, affirming the robustness and broad applicability of the BioBridge encoder. A calculated score determines the drug's binding effectiveness, the product of BioBridge's binding likelihood, and BioBridge$_r$'s binding strength, squared to emphasize the initial prediction: $ y_c^2 \times y_r $. Notably, the assessed proteins are excluded from our training data.

We randomly sampled 200 tasks from the epidermal growth factor and adenosine receptor families. We calculate the predicted scores for each task using the methods described above and tally the occurrence of query proteins in the support set for each task, as shown in Fig~\ref{fig:3}(e). For tasks with prediction scores in the top 10$\%$, accuracies excees 90$\%$. Notably, Fig~\ref{fig:3}(d) shows that the binding status of queried drug-target pairs can be accurately predicted with minimal examples, highlighting the practical utility of BioBridge in drug development with limited annotation data. In comparison to conventional inductive DTI approaches, we report the classification accuracies of BioBridge$_{Meta}$ across all tasks alongside the state-of-the-art method DrugBAN. Notably, BioBridge$_{Meta}$ achieves approximately 20$\%$ higher accuracy than DrugBAN on previously unseen targets, unequivocally demonstrating its pipeline's potential in addressing novel drug-target pairs.

\subsection{Ablation study}

To demonstrate the added value of individual modules and strategies in BioBridge, we also conduct extensive ablation studies to evaluate their effectiveness.

As shown in Table~\ref{tab:ablation_encoder}, an ablation study assesses the Multi-scale Awareness, Stem, and GAU modules of the BioBridge multi-level encoder. Removing these components shows their impacts, with BioBridge maintaining superior performance across metrics. Multi-level feature models outperformed single-level ones, highlighting their significance in knowledge transfer. As detailed in Fig~\ref{fig:cross_ablation}, The CADA module is scrutinized in cross-domain tasks, showing that BioBridge with CADA achieved the best results, followed by DrugBAN with CADA and BioBridge with CDAN. This suggests that multi-level and biological category awareness are critical to effectively capturing drug-target interactions. Moreover, the impact of cross-domain strategies on performance is significantly more significant than that of the underlying model architecture.

We also explore several key modules under the setting of the associative tasks, including Multi-levels Awareness (MA), Inductive training (LOAD), Dynamic Prototype (DGL), and Focal Loss (FL). Removing the MA module reverts it to a feature merge operation. As shown in Tables~\ref{tab:7}, the performance of the models on various few-shot tasks across the BioSNAP and BindingDB datasets decreases after removing different modules, validating the effectiveness of our design. The removal of Focal Loss also has a significant impact on performance, indicating that an adaptive focus on challenging samples is beneficial in drug discovery where the learning difficulty varies.

\section{Conclusion}
Studying protein-drug interactions is essential, particularly for proteins and drug molecules not included in training datasets. To address this challenge, we present BioBridge, a workflow that emulates chemists' induction-association approach by leveraging transferable interaction patterns and limited binding annotations to predict novel drug-target interactions. BioBridge overcomes the limitations of prior methods which struggle to balance these elements,  and improves the understanding of protein variability. Specifically, BioBridge integrates essential transferable techniques: a multi-level awareness mechanism that captures diverse interaction patterns, category-aware adversarial learning to identify transferable principles, and dynamic prototype learning to extract binding prototypes from inconsistent annotations.

As a proof-of-concept, we show that BioBridge effectively concludes transferable patterns and reliably identifies potential drug candidates, particularly for novel drug-target pairs. Extensive experiments show that BioBridge achieves a maximum 30$\%$ improvement in AUROC compared to traditional inductive methods when binding annotations are sparse. Its interpretation of multi-level binding data facilitates pharmacophore analysis and provides insights into binding sites at minimal cost. Notably, BioBridge's ability to generalize to unknown drug-target pairs using only sequence data and few binding annotations underscores its potential as a powerful tool for drug discovery. Given its theoretical scalability, we expect BioBridge's strategies to enhance future drug-target binding algorithms, improving their adaptability to new target pairs.

In this study, we present several variants of BioBridge. For well-characterized drug-target pairs, we recommend the BioBridge$_{vanilla}$. BioBridge$_{CADA}$ is the optimal choice for entirely novel targets. BioBridge$_{Meta}$ is most suitable when limited related target binding annotations are available. Computational resource requirements for each model variant are outlined in Table~\ref{tab:variantions}.

However, BioBridge has some limitations. Its multi-level encoder is slower due to hierarchical interaction extraction, and its modeling of intramolecular forces is simplistic. Additionally, it does not yet incorporate 3D protein structure data, which is limited for most proteins. However, advancements such as DeepMind's AlphaFold provide opportunities to integrate 3D structural information into future work, potentially enhancing performance and interpretability. Finally, while this study uses separate datasets, combining them with BioBridge presents an exciting avenue for future research.



\section{Methods}\label{sec3}
The core innovation of BioBridge lies in its inductive-associative framework, which mirrors a scientist's approach of combining accumulated knowledge with limited reference to tackle novel drug-target interaction predictions.

During the inductive phase, BioBridge adopts a multi-level perception strategy inspired by human reasoning. This phase further employs adversarial learning to uncover interaction principles that are transferable to previously unseen scenarios.

In the associative phase, BioBridge replicates the process of consulting related references. It uses unsupervised clustering to identify weakly related interaction annotations and enhances the model's reasoning capabilities through dynamic prototype-based meta-learning.

\subsection{BioBridge Multi-levels Encoder}

The BioBridge encoder (Figure \ref{fig:8}(a)) facilitates the analysis of protein-molecule interactions by leveraging protein sequences and molecular graphs to capture diverse interaction patterns. Inspired by human multi-level understanding and the robust transferability of shallow features in cross-domain tasks, BioBridge adopts a multi-level perceptive encoder. This design concludes interaction information incrementally, enabling broader inference of binding patterns and improved cross-domain adaptability.  Detailed architectural specifics are provided in the Section~\ref{sec:methods}.

The BioBridge encoder consists of a protein encoder $ f^p $ and a molecule encoder $ f^d $, each divided into $ I $ stages. At stage $ i $, the protein and molecule representations, $ \mathcal{P}_i $ and $ \mathcal{D}_i $, respectively, are updated using transformation functions $ f^p_i $ and $ f^d_i $:  
\begin{eqnarray}
\mathcal{P}_i = f^p_i(\mathcal{P}_{i-1}), \quad
\mathcal{D}_i = f^d_i(\mathcal{D}_{i-1})
\end{eqnarray}
This stage-wise process allows the encoder to progressively capture granular interaction patterns between proteins and molecules, reflecting their intrinsic properties.  

To model protein-molecule interactions, a binding interaction encoder $ f^b $ fuses the representations $ \mathcal{P}_i $ and $ \mathcal{D}_i $ at each stage $ i $, producing interaction features $ \mathcal{I}_i $:  
\begin{eqnarray}
\mathcal{I}_i = f^b_i(\mathcal{P}_i, \mathcal{D}_i)
\end{eqnarray} 
These features encapsulate both local and cross-level interaction patterns, offering a comprehensive representation of drug-target pairs.  

To identify the most relevant interaction patterns for drug-target interaction prediction, a selection encoder $ f^s $ integrates features across all stages. The final interaction pattern $ O $ is computed as:  
\begin{eqnarray}
O = f^s\left( \sum_{i=1}^I \mathcal{I}_i \right)
\end{eqnarray}
This process aggregates and refines the interaction features, enhancing adaptability and generalization to complex drug discovery scenarios.  

\subsection{Inductive-learning stage}\label{sec3.2}
The inductive learning phase focuses on the model's ability to generalize and transfer interaction patterns, which is assessed by its zero-shot accuracy in predicting novel drug-target interactions. Drug-target pairs are denoted as $X = \{x_1, x_2, \ldots, x_n\}$, where each pair$x_i = \{\mathcal{D}_i, \mathcal{P}_i\}$ consists of a drug $\mathcal{D}_i$ and its target protein $\mathcal{P}_i$. The corresponding interaction labels are $Y = \{y_1, y_2, \ldots, y_n\}$. The objective is to develop a model $f$ that predicts the interaction $y$ for a novel drug-target pair $x$, such that $f(x)=y$.

The multi-levels perceptron encoder mimics the comprehensive understanding scientists have of the criteria for drug-target binding. However, applying these criteria to new drug development poses challenges due to significant differences across proteins. To overcome these challenges, we propose a Category-Aware Adversarial Training (CADA) strategy during the inductive phase to identify interaction criteria applicable to novel drug-target pairs. This approach involves transferring interaction knowledge from a source domain $\text{Source} = \{(x_i, y_i)\}_{i=1}^{N_s}$, which includes $N_s$ labeled drug-protein pairs, to a target domain $\text{Target} = \{(x_i)\}_{i=1}^{N_t}$, consisting of $N_t$ unlabeled pairs.

CADA achieves this by distinguishing the domain origin of interaction representations while aligning positive and negative instances from both domains within a shared feature space. This alignment ensures that relevant interaction criteria are preserved and transferred, avoiding confusion with irrelevant patterns in the target domain.

Fig~\ref{fig:8}(d) depicts the interaction features $O_s$ and $O_t$ extracted by multi-levels aware encoder, along with predicted probabilities $\hat{y} = \{P_s^1, P_s^2, P_t^1, P_t^2\}$. A Gradient Reversal Layer (GRL) is applied to produce reversed gradients $\{R_s, R_t\}$, which are scaled by the predicted probabilities and input into the discriminator $D = \{D_0, D_1\}$, where 0 indicates negative instances, and 1 indicates positive:

\begin{eqnarray}
R_s = GRL(O_s), \quad R_t = GRL(O_t) \\   
\mathcal{L}_d(\theta_d) = \frac{1}{n} \sum_{k=1}^2 \sum_{\mathbf{x}_i \in \mathcal{R}_s \cup \mathcal{R}_t} d_i \log(D_k(\hat{y}_i^k x_i))
\end{eqnarray}

Here, $D_k$ differentiates between domains, and $d_i$ represents the domain label. The model's backbone parameters $\hat{\theta_f}$ are optimized to minimize the cross-entropy loss $\mathcal{L}_s$ on the source domain, while the discriminator's parameters $\hat{\theta_d}$ aim to maximize the adversarial loss $\mathcal{L}_d$. These competing objectives are balanced by a hyperparameter $\lambda$:

\begin{equation}
  \begin{aligned}
 \mathcal{L}(\theta_f, \theta_d) = \mathcal{L}_s(\theta_f) + \lambda \mathcal{L}_d(\theta_d)\\
 \hat{\theta_f} = \min \mathcal{L}_s(\theta_f), \quad \hat{\theta_d} = \max \mathcal{L}_d(\theta_d)
  \end{aligned}
\end{equation}

\subsection{Association-learning stage}\label{sec4_3}
We ensure that the meta-learning encoder retains transferable interaction principles learned during the inductive phase through parameter passing, enabling the expansion of the model's associative capabilities.

In the association stage, the model's ability to infer novel drug-target interactions is evaluated based on few-shot prediction accuracy. Unlike traditional methods that rely on high-quality reference data, our approach addresses real-world challenges by focusing on sparsely annotated or unknown drug-target pairs. A formal definition of meta-learning is provided in Section~\ref{secA4}.

To simulate realistic drug discovery scenarios, we extend the source and target domains by creating distinct cross-domain meta-tasks. These tasks are constructed using unsupervised clustering of broader protein categories rather than individual protein targets~\cite{wang2023zerobind}, ensuring significant divergence between training and test data. This approach reflects how researchers derive insights from weakly related references when exploring new drug-target interactions. Further details of the meta-task construction are included in Section~\ref{secA5}.

The association stage trains a meta-learning model $f_{meta}$ to predict the interaction y for a query drug-target pair Q, given a set of weakly related reference pairs $S = \{x_1, x_2, \ldots, x_n\}$. Each task comprises support sets $S = \{(X_s, Y_s)\}$ and query sets $Q = \{(X_q, Y_q)\}$. Unlike traditional gradient-based methods, our metric-based approach dynamically generates prototypes for interaction categories using learnable weights, facilitating generalization across diverse proteins.

As illustrated in Fig~\ref{fig:8}(d), the model is trained on $N$ distinct seen tasks $\mathcal{T} = \{T_1^{\text{seen}}, \ldots, T_N^{\text{seen}}\}$, organized by protein or molecule clusters. Each support set contains $k$ positive and negative samples (2-way $k$-shot). The meta-interaction encoder $f_{\text{meta}}$ derives feature representations for support interactions $\mathbf{O}_s \in \mathbb{R}^{N \times 2k \times d}$ and query interactions $\mathbf{O}_q \in \mathbb{R}^{N \times k_q \times d}$. These features are concatenated through tensor expansion operations: 
\begin{equation}
    \begin{aligned}
\mathbf{O}_s' &= \text{Expand}(\mathbf{O}_s, \text{dim}=1) \in \mathbb{R}^{N \times k_q \times 2k \times d} \\
\mathbf{O}_c &= [\mathbf{O}_q \otimes \mathbf{1}_{2k}; \mathbf{O}_s'] \in \mathbb{R}^{N \times k_q \times (2k+1) \times d}
\end{aligned}
\end{equation}
where $\text{Expand}(\cdot)$ denotes dimension expansion with replication, $[\cdot; \cdot]$ represents concatenation along the third dimension, and $\otimes$ indicates tensor broadcasting. The class prototypes $\mathbf{P}$ are computed through attention mechanisms:
\begin{equation}
 \begin{aligned}
\mathbf{Q} &= \sigma(\mathbf{W}_I\mathbf{O}_c) \circ \gamma_1 + \beta_1 \\
\mathbf{K} &= \sigma(\mathbf{W}_I\mathbf{O}_c) \circ \gamma_2 + \beta_2 \\
\mathbf{A} &= \left[\text{ReLU}(\mathbf{Q}\mathbf{K}^\top)\right]^{\odot 2}
    \end{aligned}
\end{equation}
\begin{equation}
    \begin{aligned}
\mathbf{P}[n, q, c] = \frac{1}{|\mathcal{O}_c|} \sum_{i\in\mathcal{O}_c} \mathbf{O}_s'[n, q, i, :] \cdot \frac{\exp(\mathbf{A}[n,q,i])}{\sum_{j=1}^{2k}\exp(\mathbf{A}[n,q,j])}
\end{aligned}
\end{equation}

where $\sigma$ denotes the SiLU activation function, $\circ$ represents element-wise multiplication, and $\odot 2$ indicates element-wise square.

To classify $ X_q $, we compute the cosine similarity between $ P $ and $ X_q $. Recognizing that negative samples are easier to classify in drug-target interactions, we use Focal Loss to emphasize harder-to-classify cases. This dynamically reduces the weight of easily distinguishable samples, focusing on challenging ones: 
\begin{equation}
    \begin{aligned}s_{nqc} &= \cos(\mathbf{P}[n,q,c], \mathbf{X}_q[n,q]) \\
p_{nq} &= \sum_{c\in\{0,1\}} \frac{\exp(s_{nqc})}{\sum_{c'}\exp(s_{nqc'})} \cdot \mathbb{I}(Y_q = c) \\
\mathcal{L}_f &= -\alpha \sum_{n,q} (1-p_{nq})^\gamma \log(p_{nq})
\end{aligned}
\end{equation}

Here, $ \mathbb{I}(\cdot)$ is the Kronecker delta function, which equals 1 if $ Y_q = c $ and 0 otherwise. The class weight $ \alpha $ balances losses between positive and negative samples, while the modulation factor $ (1-p_{nq})^\gamma $ minimizes the impact of easily classified samples and emphasizes challenging ones. The final parameters are obtained by $\hat{\theta}_m = \arg\min_{\theta_m} \mathcal{L}_f$.

\section{Data availability}
The experimental data with each split strategy are available at https://github.com/lian-xiao/BioBridge .

\section{Code availability}
The source code and implementation details of BioBridge are freely available at https://github.com/lian-xiao/BioBridge .

\section{Acknowledgement}
This work is supported in part by the National Natural Science Foundation of China under grants W2411054, U21A20521 and 62271178, the Postgraduate Research \& Practice Innovation Program of Jiangsu Province KYCX23\_2524, National Foreign Expert Project of China under Grant G2023144009L, Zhejiang Provincial Natural Science Foundation of China (LR23F010002), Wuxi Health Commission Precision Medicine Project (J202106), Jiangsu Provincial Six Talent Peaks Project (YY-124), and the construction project of Shanghai Key Laboratory of Molecular Imaging (18DZ2260400).

\bibliography{sn-bibliography}

\begin{appendices}
\onecolumn
\setcounter{table}{0}  
\setcounter{figure}{0}
\renewcommand{\thetable}{A\arabic{table}}
\renewcommand{\thefigure}{A\arabic{figure}}

\section{Supplementary Material}\label{secA1}
\subsection{Datasets}
We utilize three key datasets for Drug-Target Interaction (DTI) research: BindingDB, BioSNAP, and Human. BindingDB is an online database that provides experimentally confirmed binding affinities between small molecules and proteins. It is known for its reliability. We employ a low-bias version developed by Bai~\cite{bai2021hierarchical}. The BioSNAP dataset, compiled from DrugBank by Huang~\cite{huang2021moltrans} and Marinka~\cite{zitnik2018biosnap}, includes data on nearly 5,000 drugs and over 2,000 proteins. It is characterized by a balanced distribution of positive instances (known interactions) and randomly selected negative instances to ensure dataset equilibrium. The Human dataset, assembled by Liu~\cite{liu2015improving}, is noted for its high-quality negative samples derived from advanced computational screening methods. We use a balanced version of this dataset, containing an equal number of positive and negative samples. Additional details about these datasets are provided in Table~\ref{tab:app:datasets}.


\begin{table*}[!ht]
    \centering
    \rowcolors*{1}{myco}{white}
    \caption{Experimental dataset statistics.}
    \phantomsection
    \label{tab:app:datasets}
    \begin{tabular}{cccc}
    \hline
        Dataset & Drugs & Targets & Interactions \\ \hline
        BingdingDB~\cite{bai2021hierarchical} & 14643 & 2623 & 49199 \\ 
        BioSNAP~\cite{hua2022transformer} & 4510 & 2181 & 27464 \\ 
        Human~\cite{liu2015improving} & 2726 & 2001 & 6728 \\ \hline
    \end{tabular}
    
\end{table*}

\subsection{Notations and descriptions}\label{secA2}
Table~\ref{tab:variantions} represents variations and explanations of BioBridge.
\begin{table*}[!ht]
\centering
\rowcolors*{1}{myco}{white}
\caption{Variations and explanations of BioBridge.}
\phantomsection
\label{tab:variantions}
\begin{tabularx}{\textwidth}{cX}
    \hline
    \textbf{Model Name} & \textbf{Explanation} \\
    \hline
    BioBridge$_{vanilla}$ & The BioBridge model without the Category-Aware Domain Adversarial Learning (CADA) module. \\
    BioBridge$_{CDAN}$ & The model that integrates the CDAN module lacking category domain awareness into DrugBAN or BioBridge, used for comparative analysis. \\
    BioBridge$_{CADA}$ & The BioBridge model combined with the Category-Aware Domain Adversarial Learning (CADA) module, which is used for identifying interaction patterns applicable to unknown drug-target pairs and performs well in cross-domain DTI prediction. \\
    BioBridge$_{Meta}$ & The BioBridge model used in the meta-learning setting, which predicts novel drug-protein interactions by extracting transferable interaction prototypes from the support set with limited data. \\
    \hline
\end{tabularx}
\label{tab:a2}
\end{table*}

\subsection{Experimental setting}\label{secA3}

\textbf{Evaluation strategies and metrics: }
We evaluate our model's classification accuracy across three public datasets: BindingDB~\cite{gilson2016bindingdb,bai2021hierarchical}, BioSNAP~\cite{zitnik2018biosnap,hua2022transformer}, and Human~\cite{liu2015improving}, with the test set simulating an unknown real-world scenario. Table~\ref{tab:app:datasets} outlines dataset characteristics. Considering the bias highlighted by Chen et al.~\cite{chen2020transformercpi}, which suggests that models tend to memorize existing drug patterns rather than actual interactions, we primarily assess the model's generalization capabilities on unknown drug-target pairs. We investigate two data partitioning strategies: the cold pair split, which allocates 70$\%$ of drug/protein pairs for training and splits the remaining 30$\%$ into validation (30$\%$) and test sets (70$\%$), and the cross-domain partitioning, which follows the clustering-based approach of DrugBAN~\cite{bai2023interpretable}. This cold pair split strategy guarantees that all test drugs and proteins are not observed during training so that prediction on test data cannot rely only on the features of known drugs or protein. For the latter, drugs and proteins are clustered using ECFP4 and PSC algorithms, respectively, with 60$\%$ of clusters designated as the source domain and the remaining 40$\%$ as the target domain. The cluster-based partitioning ensures that the model does not even see drug-target pairs similar to those in the test set.

In meta-learning contexts, we apply PSC clustering to proteins in BindingDB and BioSNAP datasets, coupled with scaffold splitting for drug molecules. We define tasks based on protein and drug molecule clusters to ensure diverse yet related query and support sets, mirroring the drug discovery scenario with limited novel drug-target pairs but available informative interactions. Tasks with fewer than six instances are consolidated into the source domain, enriching its diversity and preparing for a challenging target domain task. The first 40$\%$ of cumulative distribution clusters are allocated to the source domain, with the rest forming the target domain, further divided into training (70$\%$) and test sets (30$\%$) to mimic real-world distribution variations. Testing incorporates at least twice the number of tasks as the test set size, establishing a stringent training and evaluation framework. Clustering details are detailed in Table~\ref{tab:a3}.

We use AUROC (area under the receiver operating characteristic curve), AUPRC (area under the precision-recall curve), and ACC (Accuracy) as primary metrics for classification tasks. The model with the highest AUROC on the validation set is selected for testing, with results reported on the test set. To ensure reliability, we conduct five independent runs with different random seeds.

\textbf{Implementation: }
For conventional training, the BioBridge model utilizes the Adam optimizer at a learning rate 5e-5 with batch sizes of 64 over 100 epochs. In contrast, for meta-learning scenarios, the learning rate is increased to 1e-4, with batch sizes reduced to 32, and the training is conducted for 50 epochs. The best-performing model is selected at the epoch, giving the best AUROC score on the validation set, which is then used to evaluate the final performance of the test set. The protein feature stem consists of two 128-dimensional 1D convolutional layers, and the molecule feature stem has two fully connected layers with 128 dimensions. The target feature encoder has three 1D-CNN layers with 128 filters and kernel sizes of 3, 6, and 9, along with max-pooling layers. The drug feature encoder includes three GCN layers with 128 dimensions for feature selection. Max-pooling and GCN layers with 128 dimensions perform feature selection. The maximum sequence length for targets is 1200, and the maximum number of atoms for drug molecules is 290. The bilinear attention module has two heads with a latent embedding size of 768 and an average pooling window size of 3. The gated attention unit has a total hidden layer size of 256, with Query and Key layers at 128. The fully connected decoder has 512 hidden neurons. The implementation of meta-learning is based on learn2learn~\cite{arnold2020learn2learn}.

\textbf{Baselines: }
In summary, we compare our model with seven others for villain DTI prediction: (1) GNN-CPI~\cite{tsubaki2019compound}, which employs a graph neural network for drugs and a CNN for proteins, linking latent vectors for interaction prediction; (2) DeepConv-DTI~\cite{lee2019deepconv}, which utilizes a CNN with global max-pooling for protein sequences and a fully connected network for ECFP4 drug fingerprints; (3) GraphDTA~\cite{nguyen2021graphdta}, which combines a graph neural network for drug molecular graphs with a CNN for protein sequences through simple concatenation; (4) TransformerCPI~\cite{chen2020transformercpi}, which enhances interaction prediction using sequence-based deep learning with a self-attention mechanism; (5) MolTrans~\cite{huang2021moltrans}, which applies a Transformer for encoding and a CNN-based module for substructure interactions; and (6) DrugBAN~\cite{bai2023interpretable}, which models DTI using a graph neural network for drug molecular graphs and a CNN for proteins, incorporating a bilinear attention mechanism.

Given the innovative nature of our research, existing meta-learning studies focused on specialized domains and tailored encoders offer limited applicability. Therefore, we compare our model against broader meta-learning approaches that may yield more versatile solutions. In our meta-task experiments, we use BioBridge as the primary network for data encoding, evaluating its performance against five advanced learning strategies designed for rapid adaptation: (1) MAML++~\cite{MAMLPP}, which enhances MAML~\cite{finn2017model} by introducing an additional task for improved adaptability; (2) ANIL~\cite{raghu2019rapid}, which employs a separate network to predict necessary model adjustments for new tasks; (3) Prototypical Networks~\cite{snell2017prototypical}, which learns from representative examples to quickly grasp new tasks; and (4) MetaOptNet~\cite{lee2019meta}, which refines model performance on new tasks by fine-tuning essential hyperparameters.

We follow the recommended hyperparameter settings for each model.

\subsection{Meta-learning problem definition}\label{secA4}
The proposed meta-learning framework identifies shared patterns across tasks and adapts quickly to new tasks, improving predictive accuracy. It is framed as an $N_w$-way, $N_s$-shot problem, where $N_w$ is the number of classes, and $N_s$ is the number of labeled examples per class. Each training iteration samples $N_w$ classes to construct a task $T$. Drug-target pairs are represented as $X = \{x_1, x_2, \ldots, x_n\}$, where $x_i = \{\mathcal{D}_i, \mathcal{P}_i\}$ includes a drug $\mathcal{D}_i$ and its target protein $\mathcal{P}_i$. Corresponding class labels are $Y = \{y_1, y_2, \ldots, y_n\}$. Task $T$ includes a support set $S = \{(X_s, Y_s)\}$ and a query set $Q = \{(X_q, Y_q)\}$, constructed by randomly selecting $N_s$ and $N_q$ samples per class. The goal is to train the model to adapt to the query set $Q$ using the support set $S$. In drug-target interaction (DTI) prediction, $N_w = 2$.

\subsection{Meta unseen pair split strategy}\label{secA5}
For cross-domain meta-learning performance evaluation, we perform clustering on target proteins from the BindingDB and BioSNAP datasets using single-linkage clustering. This method maintains inter-cluster distances above a threshold $\lambda$. Furthermore, molecular scaffold-based clustering is applied to the molecules within these datasets. This method fosters a cross-domain environment by avoiding cluster convergence and preserving intra-cluster homogeneity, mirroring the drug discovery process where existing drug-target interactions inform the prediction of new ones. 

We utilize integral PSC features to represent target proteins and measure pairwise distances using Jaccard distance for ECFP4 and cosine distance for PSC. We set $\lambda = 0.5$ in protein clustering to prevent large cluster formation and ensure sample separation. Table~\ref{tab:cluster_details} presents the sample counts of the ten most significant clusters, revealing that BindingDB has a more uniform cluster distribution than BioSNAP in drug clustering. Additionally, protein clustering tends to form many small clusters across both datasets, indicating lower average similarity among proteins than drugs.

\begin{table*}[h]
    \centering
    \rowcolors*{1}{myco}{white}
    \caption{Size of the ten largest clusters in the BindingDB and BioSNAP datasets generated by the meta unseen pair split.}
    \phantomsection
    \label{tab:cluster_details}
    \begin{tabular}{cccccccccccc}
        \Xhline{1.2pt}
    \textbf{Dataset} & \textbf{Object} & \textbf{\#1} & \textbf{\#2} & \textbf{\#3} & \textbf{\#4} & \textbf{\#5} & \textbf{\#6} & \textbf{\#7} & \textbf{\#8} & \textbf{\#9} & \textbf{\#10} \\
    \Xhline{1.2pt}
    BindingDB        & Drug            & 175          & 116          & 105          & 90           & 82           & 75           & 74           & 63           & 56           & 53            \\
    BioSNAP          & Drug            & 475          & 353          & 59           & 56           & 37           & 32           & 29           & 26           & 23           & 22            \\
    BindingDB        & Protein         & 17           & 15           & 15           & 12           & 10           & 10           & 10           & 9            & 9            & 8             \\
    BioSNAP          & Protein         & 8            & 8            & 8            & 6            & 5            & 4            & 4            & 4            & 4            & 4    \\
    \Xhline{1.2pt}        
    \end{tabular}
    
\end{table*}

\subsection{Scalability and computational efficiency}\label{secA6}
\begin{figure}[!ht]
    \centering
    \includegraphics[width=\linewidth]{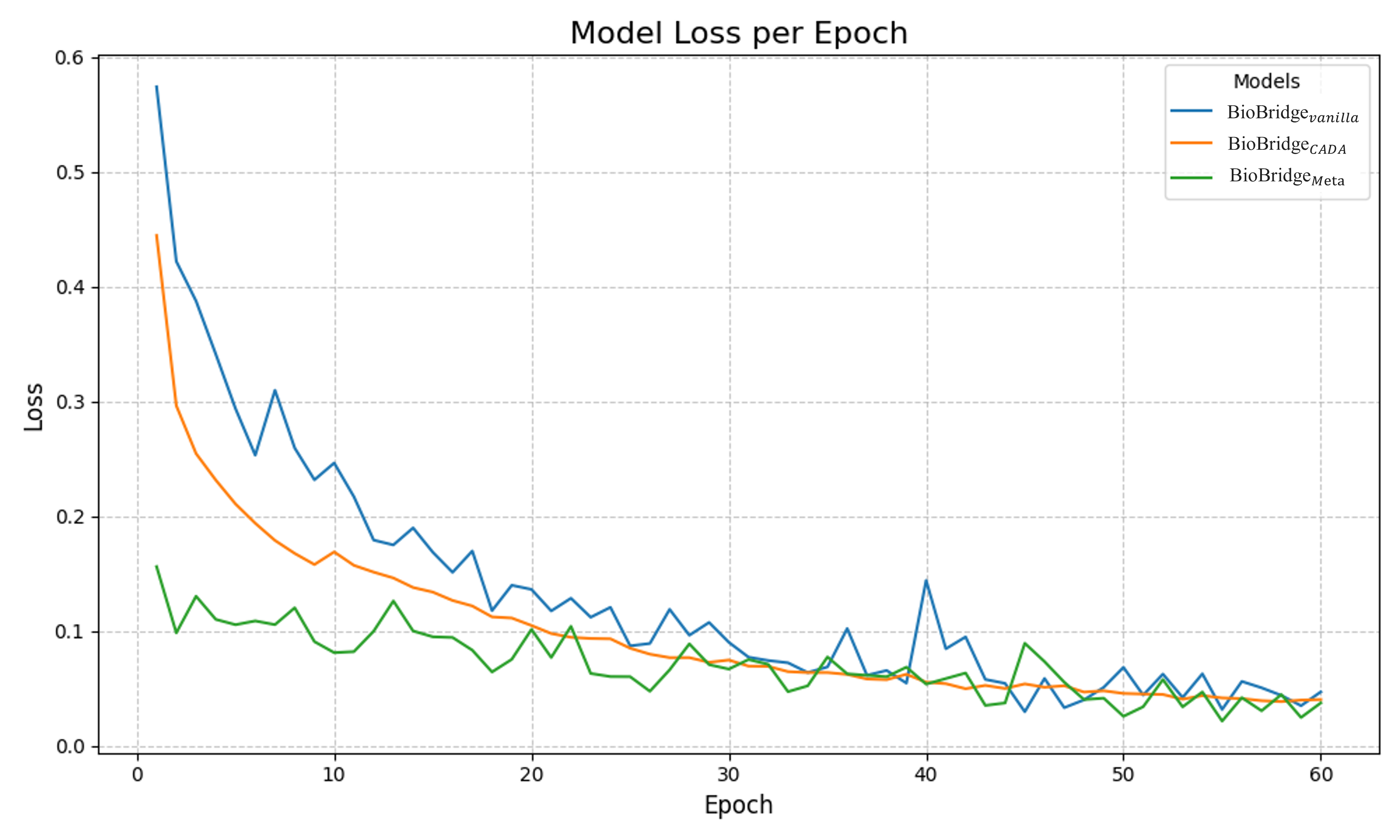}
    \caption{The training losses of the different variants of BioBridge vary with epoch.}
    \phantomsection
    \label{fig:loss}
\end{figure}

Designed for scalability, BioBridge achieves optimal performance through component integration. As shown in Fig~\ref{fig:loss},BioBrigde demands minimal training resources, fitting well within 60 epochs on an RTX4090 and can be expedited with early stopping. At a batch size of 64, the vanilla BioBridge completes an epoch in 1.53 minutes using 9.2G of VRAM. BioBridge$_{CADA}$ takes 2.5 minutes per epoch with 13.7G VRAM, while BioBridge$_{Meta}$ requires just 35 seconds per epoch with 2.1G VRAM.

\subsection{Comparison on random and cold-pair split}
Table~\ref{tab:a4} compares the performance of BioBridge on random versus cold splits. Unlike on cold splits, BioBridge's advantage on random splits is less pronounced, suggesting that its multi-level interaction patterns are better suited for unknown drug target pairs, while still adequately addressing known pairs.

\begin{table*}[h]
    \renewcommand{\arraystretch}{1.2}
    \centering
    \rowcolors*{1}{myco}{white}
    \caption{Performance comparison on the BindingDB, BioSNAP  and Human datasets(\textbf{Best},\uline{ Second Best}).}
    \phantomsection
    \label{tab:a4}
    \resizebox{\linewidth}{!}{
        \begin{tabular}{ccccccc}
            \Xhline{1.2pt}
            Metrics                                      & AUROC                    & AUPRC                    & ACC                      & AUROC                    & AUPRC                    & ACC                      \\
            \Xhline{1.2pt}
            \rowcolor{myco2} 
            Datasets                    & \multicolumn{3}{c}{BindingDB}                                                  & \multicolumn{3}{c}{BindingDB(Cold)}                                            \\

            GNN-CPI~\cite{tsubaki2019compound} & $ 0.9418\pm 0.0051^* $ & $ 0.9210\pm 0.0113^* $ & $ 0.8823\pm 0.0021^* $ & $ 0.5893\pm 0.0101^* $ & $ 0.5562\pm 0.0140^* $ & $ 0.5481\pm 0.0182^* $ \\
            DeepConv-DTI~\cite{lee2019deepconv}  & $ 0.9502\pm 0.0030^* $ & $ 0.9308\pm 0.0042^* $ & $ 0.8924\pm 0.0027^* $ & $ 0.6204\pm 0.0128^* $ & $ 0.5913\pm 0.0154^* $ & $ 0.5879\pm 0.0251^* $ \\
            GraphDTA~\cite{nguyen2021graphdta}  & $ 0.9517\pm 0.0009^* $ & $ 0.9371\pm 0.0009^* $ & $ 0.8989\pm 0.0023^* $ & $ 0.6191\pm 0.0142^* $ & $ \uline{0.6006\pm 0.0201} $ & $ 0.6003\pm 0.0205^* $ \\
            TransformerCPI~\cite{chen2020transformercpi} & $ 0.9511\pm 0.0012^* $ & $ 0.9489\pm 0.0017^* $ & $ 0.8954\pm 0.0011^* $ & $ 0.6559\pm 0.0124^* $ & $ 0.5941\pm 0.0132^* $ & $ 0.6096\pm 0.0131^* $ \\
            HyperAttDTI~\cite{zhao2022hyperattentiondti} & $ 0.9593\pm 0.0017^* $ & $ \uline{0.9513\pm 0.0018^*} $ & $ 0.9012\pm 0.0028^* $ & $ \uline{0.6612\pm 0.0161} $ & $ 0.5988\pm 0.0213 $ & $ \uline{0.6332\pm 0.0226^*} $ \\
            MolTrans~\cite{huang2021moltrans} & $ 0.9541\pm 0.0022^* $ & $ 0.9414\pm 0.0009^* $ & $ 0.9023\pm 0.0012^* $ & $ 0.5948\pm 0.0144^* $ & $ 0.5225\pm 0.0176^* $ & $ 0.5324\pm 0.0148^* $ \\
            DrugBAN~\cite{bai2023interpretable}  & $ \uline{0.9624\pm 0.0009^*} $ & $ 0.9512\pm 0.0009^* $ & $ \uline{0.9064\pm 0.0041^*} $ & $ 0.6553\pm 0.0187^* $ & $ 0.6004\pm 0.0151 $ & $ 0.6121\pm 0.0212^* $ \\
            BioBridge & $ \bm{0.9662\pm 0.0009} $ & $ \bm{0.9543\pm 0.0009} $ & $ \bm{0.9143\pm 0.0032} $ & $ \bm{0.6801\pm 0.0151} $ & $ \bm{0.6269\pm 0.0197} $ & $ \bm{0.6341\pm 0.0271} $ \\
            \rowcolor{myco2} Datasets                    & \multicolumn{3}{c}{BioSNAP}                                                    & \multicolumn{3}{c}{BioSANP(Cold)}                                              \\
            GNN-CPI~\cite{tsubaki2019compound} & $ 0.8832\pm 0.0052^* $ & $ 0.8913\pm 0.0042^* $ & $ 0.8171\pm 0.0024^* $ & $ 0.6331\pm 0.0153^* $ & $ 0.6574\pm 0.0112^* $ & $ 0.6162\pm 0.0091^* $ \\
            DeepConv-DTI~\cite{lee2019deepconv} & $ 0.8900\pm 0.0041^* $ & $ 0.8954\pm 0.0036^* $ & $ 0.8152\pm 0.0053^* $ & $ 0.6534\pm 0.0104^* $ & $ 0.6683\pm 0.0121^* $ & $ 0.6187\pm 0.0136^* $ \\
            GraphDTA~\cite{nguyen2021graphdta} & $ 0.8901\pm 0.0082^* $ & $ 0.8921\pm 0.0071^* $ & $ 0.8103\pm 0.0067^* $ & $ 0.6431\pm 0.0112^* $ & $ 0.6730\pm 0.0231^* $ & $ 0.6315\pm 0.0093^* $ \\
            TransformerCPI~\cite{chen2020transformercpi} & $ 0.8945\pm 0.0016^* $ & $ 0.8931\pm 0.0033^* $ & $ 0.8225\pm 0.0045^* $ & $ 0.6803\pm 0.0099^* $ & $ 0.7083\pm 0.0079^* $ & $ 0.6483\pm 0.0119 $ \\
            HyperAttDTI~\cite{zhao2022hyperattentiondti} & $ 0.9042\pm 0.0029^* $ & $ 0.9087\pm 0.0039^* $ & $ 0.8312\pm 0.0040^* $ & $ \uline{0.7148\pm 0.0182} $ & $ \bm{0.7312\pm 0.0150} $ & $ \uline{0.6636\pm 0.0172} $ \\
            MolTrans~\cite{huang2021moltrans} & $ 0.9019\pm 0.0037^* $ & $ 0.9044\pm 0.0047^* $ & $ 0.8254\pm 0.0102^* $ & $ 0.6724\pm 0.0146^* $ & $ 0.6968\pm 0.0104^* $ & $ 0.6402\pm 0.0118 $ \\
            DrugBAN~\cite{bai2023interpretable} & $ \uline{0.9084\pm 0.0022^*} $ & $ \uline{0.9119\pm 0.0041^*} $ & $ \uline{0.8339\pm 0.0083^*} $ & $ 0.6589\pm 0.0191^* $ & $ 0.6673\pm 0.0119^* $ & $ 0.6324\pm 0.0122^* $ \\
            BioBridge & $ \bm{0.9161\pm 0.0021} $ & $ \bm{0.9192\pm 0.0031} $ & $ \bm{0.8428\pm 0.0018} $ & $ \bm{0.7161\pm 0.0110} $ & $ \uline{0.7231\pm 0.0108} $ & $ \bm{0.6648\pm 0.0187} $ \\
            \rowcolor{myco2} 
            Datasets                    & \multicolumn{3}{c}{Human}                                                      & \multicolumn{3}{c}{Human(Cold)}                                                \\
            GNN-CPI~\cite{tsubaki2019compound} & $ 0.9791\pm 0.0034^* $ & $ 0.9807\pm 0.0031 $ & $ 0.9191\pm 0.0100 $ & $ 0.8143\pm 0.0231^* $ & $ 0.7514\pm 0.0308 $ & $ 0.7047\pm 0.0311^* $ \\
            DeepConv-DTI~\cite{lee2019deepconv} & $ 0.9802\pm 0.0021^* $ & $ 0.9813\pm 0.0013 $ & $ 0.9200\pm 0.0037 $ & $ 0.8317\pm 0.0452^* $ & $ 0.7856\pm 0.0524 $ & $ 0.7281\pm 0.0182^* $ \\
            GraphDTA~\cite{nguyen2021graphdta} & $ 0.9809\pm 0.0012^* $ & $ \bm{0.9822\pm 0.0009} $ & $ 0.9242\pm 0.0056 $ & $ 0.8301\pm 0.0391 $ & $ 0.7723\pm 0.0463 $ & $ 0.7192\pm 0.0353 $ \\
            TransformerCPI~\cite{chen2020transformercpi} & $ 0.9730\pm 0.0011^* $ & $ 0.9756\pm 0.0021^* $ & $ \uline{0.9358\pm 0.0101} $ & $ 0.8382\pm 0.0233^* $ & $ 0.7712\pm 0.0270 $ & $ 0.7314\pm 0.0196^* $ \\
            HyperAttDTI~\cite{zhao2022hyperattentiondti} & $ \bm{0.9842\pm 0.0011} $ & $ 0.9812\pm 0.0029 $ & $ 0.9354\pm 0.0071 $ & $ 0.8489\pm 0.0321 $ & $ 0.7870\pm 0.0271 $ & $ 0.7563\pm 0.0287 $ \\
            MolTrans~\cite{huang2021moltrans} & $ 0.9798\pm 0.0022^* $ & $ 0.9791\pm 0.0018 $ & $ 0.9322\pm 0.0119 $ & $ 0.8342\pm 0.0257^* $ & $ 0.7821\pm 0.0384 $ & $ 0.7520\pm 0.0194 $ \\
            DrugBAN~\cite{bai2023interpretable} & $ 0.9823\pm 0.0021 $ & $ 0.9819\pm 0.0031 $ & $ 0.9334\pm 0.0088 $ & $ \uline{0.8511\pm 0.0276} $ & $ \uline{0.7894\pm 0.0427} $ & $ \uline{0.7574\pm 0.0311} $ \\
            BioBridge & $ \uline{0.9842\pm 0.0031} $ & $ \uline{0.9822\pm 0.0027} $ & $ \bm{0.9359\pm 0.0102} $ & $ \bm{0.8772\pm 0.0341} $ & $ \bm{0.7951\pm 0.0582} $ & $ \bm{0.7618\pm 0.0240} $ \\
            \Xhline{1.2pt} 
            \rowcolor{white} \multicolumn{7}{l}{\small $*$ Significantly different (p < 0.05) from the corresponding BioBridge metric value; one-way analysis of variance (ANOVA).}\\
        \end{tabular}
    }
    \label{tab:a4}
\end{table*}

\subsection{Compared with the structure-based protein ligand model}
We compare BioBridge against structure-based protein-ligand models and find it to be the best among models not requiring 3D crystal structures, matching some structure-based models in performance. This suggests that BioBridge can be used cost-effectively for accurate drug target binding affinity predictions and provides interpretable binding site insights.

\begin{table*}[!ht]
    \renewcommand{\arraystretch}{1.2}
    \centering
    \rowcolors*{1}{myco}{white}
    \caption{ Performance comparison on the PDB2020 datasets. (\textbf{Best Sequence-based},\uline{Second Best Sequence-based}).}
    \phantomsection
    \label{tab:a5}
    \resizebox{\linewidth}{!}{
        \begin{tabular}{ccccccccc}
            \Xhline{1.2pt} 
            Methods   & Hi-res                       & Co-crystal  & \multicolumn{6}{c}{PDBBind v2020}                                                                                    \\
                      & Structure & Complex      & RMSE $\downarrow$ & MAE $\downarrow$ & Pearson $\uparrow$ & Spearman $\uparrow$ & $r_m^2 \uparrow$ & CI $\uparrow$   \\ \Xhline{1.2pt} 
                      Pafnucy~\cite{stepniewska2018development} & \ding{51} & \ding{51} &$ 1.4353\pm 0.0182 $ & $ 1.1447\pm 0.0189^* $ & $ 0.6359\pm 0.0083^* $ & $ 0.5872\pm 0.0089^* $ & $ 0.3487\pm 0.0162 $ & $ 0.7074\pm 0.0041^* $ \\
                      OnionNet~\cite{zheng2019onionnet} & \ding{51} & \ding{51} &$ 1.4039\pm 0.0124 $ & $ 1.1038\pm 0.0147 $ & $ 0.6483\pm 0.0079 $ & $ 0.6024\pm 0.0132^* $ & $ 0.3813\pm 0.0119^* $ & $ 0.7178\pm 0.0053^* $ \\
                      IGN~\cite{jiang2021interactiongraphnet} & \ding{51} & \ding{51} & $ 1.4047\pm 0.0253 $ & $ 1.1169\pm 0.0304 $ & $ 0.6623\pm 0.0137 $ & $ 0.6389\pm 0.0214 $ & $ 0.3854\pm 0.0209^* $ & $ 0.7309\pm 0.0094 $ \\
                      SMINA~\cite{koes2013lessons} & \ding{51} & \ding{55} &$ 1.4662\pm 0.0089^* $ & $ 1.1613\pm 0.0074^* $ & $ 0.6658\pm 0.0057 $ & $ 0.6639\pm 0.0192^* $ & $ 0.3917\pm 0.0313 $& $ 0.7408\pm 0.0082 $ \\
                      GNNA~\cite{mcnutt2021gnina} & \ding{51} & \ding{55} &$ 1.7409\pm 0.0143^* $ & $ 1.4137\pm 0.0152^* $ & $ 0.4958\pm 0.0119^* $ & $ 0.4947\pm 0.0113^* $ & $ 0.2094\pm 0.0098^* $ & $ 0.6749\pm 0.0047^* $ \\
                      dMaSIF~\cite{sverrisson2021fast} & \ding{51} & \ding{55} &$ 1.4509\pm 0.0327 $ & $ 1.1364\pm 0.0319 $ & $ 0.6298\pm 0.0183^* $ & $ 0.5889\pm 0.0412 $ & $ 0.3479\pm 0.0297 $ & $ 0.7107\pm 0.0173 $ \\
                      GraphDTA~\cite{nguyen2021graphdta} & \ding{55} & \ding{55} & $ 1.5640\pm 0.0630^* $ & $1.2230\pm 0.0660^* $ & $ 0.6120\pm 0.0160^* $ & $ 0.5700\pm 0.0500^* $ & $ 0.3060\pm 0.0390^* $ & $ 0.7030\pm 0.0190^* $ \\
                      TransformerCPI~\cite{chen2020transformercpi}  & \ding{55} & \ding{55} & $ 1.4930\pm 0.0500^* $ & $ 1.2010\pm 0.0370^* $ & $ 0.6040\pm 0.0240^* $ & $ 0.5510\pm 0.0290^* $ & $ 0.2550\pm 0.0270^* $ & $ 0.6770\pm 0.0110^* $ \\
                      MolTrans~\cite{huang2021moltrans} & \ding{55} & \ding{55} & $ 1.5990\pm 0.0600^* $ & $ 1.2710\pm 0.0510^* $ & $ 0.5390\pm 0.0570^* $ & $ 0.4740\pm 0.0520^* $ & $ 0.2420\pm 0.0450^* $ & $ 0.6660\pm 0.0200^* $ \\
                      DrugBAN~\cite{bai2023interpretable} & \ding{55} & \ding{55} & $ \uline{1.4800\pm 0.0460^*} $ & $ \uline{1.1590\pm 0.0450^*} $ & $ \uline{0.6570\pm 0.0180} $ & $ \uline{0.6120\pm 0.0270^*} $ & $ \uline{ 0.3190\pm 0.0210^* } $ & $ \uline{0.7200\pm 0.0110}$ \\
                      BioBridge & \ding{55} & \ding{55} &$ \bm{1.4172\pm 0.0012 }$ & $ \bm{1.1038\pm 0.0139} $ & $ \bm{0.6636\pm 0.0090} $ & $ \bm{0.6331\pm 0.0170} $ & $ \bm{0.3543\pm 0.0180} $ & $ \bm{0.7285\pm 0.0090} $ \\
                      \Xhline{1.2pt}
                      \rowcolor{white} \multicolumn{7}{l}{\small $*$ Significantly different (p < 0.05) from the corresponding BioBridge metric value; one-way analysis of variance (ANOVA).}\\
            \end{tabular}
    }
    \label{tab:a5}
\end{table*}

\subsection{Comparison on specific meta unseen protein split}
Under the specific meta-unseen protein split paradigm, each meta-task's support and query sets comprise binding profiles from the same target with distinct drugs. This stringent meta-learning strategy necessitates predicting new interactions with consistent drug-target pairs.
\begin{table*}[!ht]
    \renewcommand{\arraystretch}{1.2}
    \centering
    \rowcolors*{1}{myco}{white}
    \caption{Few shot comparison of specific meta unseen protein splitting on the BindingDB and BinSNAP datasets (\textbf{Best},\uline{Second Best}).}
    \phantomsection
    \label{tab:3}
    \resizebox{\linewidth}{!}{
    \begin{tabular}{ccccccc}
    \Xhline{1.2pt} Dataset                  & \multicolumn{3}{c}{BindingDB(Specific Meta Unseen Protein)}                             & \multicolumn{3}{c}{BioSNAP(Specific Meta Unseen Protein)}                               \\
    Metric                                  & AUROC                       & AUPRC                       & ACC                         & AUROC                       & AUPRC                       & ACC                         \\
    \Xhline{1.2pt} 
    \rowcolor{myco2} Setting & \multicolumn{6}{c}{1-shot}                                                                                                                                                        \\
    MAML++~\cite{MAMLPP} & $ 0.6477\pm 0.0133^* $ & $ 0.6343\pm 0.0164^* $ & $ 0.6061\pm 0.0093^* $ & $ 0.5804\pm 0.0108^* $ & $ 0.5705\pm 0.0125^* $ & $ 0.5565\pm 0.0082^* $ \\
    Protypes~\cite{snell2017prototypical} & $ 0.7186\pm 0.0129^* $ & $ \uline{0.7209\pm 0.0154^*} $ & $ 0.6500\pm 0.0087^* $ & $ 0.6601\pm 0.0049^* $ & $ 0.6543\pm 0.0055^* $ & $ 0.6113\pm 0.0067^* $ \\
    MetaOptNet~\cite{lee2019meta} & $ \uline{0.7344\pm 0.0062^*} $ & $ 0.7191\pm 0.0063^* $ & $ \uline{0.6711\pm 0.0036^*} $ & $ \uline{0.6683\pm 0.0097^*} $ & $ 0.6469\pm 0.0099^* $ & $ \uline{0.6248\pm 0.0046} $ \\
    ANIL~\cite{raghu2019rapid} & $ 0.6951\pm 0.0247^* $ & $ 0.6918\pm 0.0259^* $ & $ 0.6374\pm 0.0181^* $ & $ 0.6560\pm 0.0280^* $ & $ \uline{0.6548\pm 0.0214} $ & $ 0.6051\pm 0.0239 $ \\
    BioBridge$_{Meta}$ & $ \bm{0.7824\pm 0.0047} $ & $ \bm{0.7893\pm 0.0050} $ & $ \bm{0.6960\pm 0.0027} $ & $ \bm{0.6861\pm 0.0128} $ & $ \bm{0.6841\pm 0.0139} $ & $ \bm{0.6286\pm 0.0091} $ \\
    \rowcolor{myco2} Setting                & \multicolumn{6}{c}{3-shot}                                                                                                                                                        \\
    MAML++~\cite{MAMLPP} & $ 0.7831\pm 0.0156^* $ & $ 0.7806\pm 0.0160^* $ & $ 0.7079\pm 0.0152^* $ & $ 0.6783\pm 0.0026^* $ & $ 0.6723\pm 0.0017^* $ & $ 0.6257\pm 0.0032^* $ \\
    Protypes~\cite{snell2017prototypical} & $ 0.7582\pm 0.0207^* $ & $ 0.7588\pm 0.0208^* $ & $ 0.6849\pm 0.0195^* $ & $ 0.7160\pm 0.0107^* $ & $ 0.7103\pm 0.0111^* $ & $ 0.6541\pm 0.0032^* $ \\
    MetaOptNet~\cite{lee2019meta} & $ \uline{0.8311\pm 0.0070^*} $ & $ \uline{0.8114\pm 0.0048^*} $ & $ \uline{0.7604\pm 0.0083} $ & $ \uline{0.7568\pm 0.0077^*} $ & $ \uline{0.7341\pm 0.0061^*} $ & $ \uline{0.6967\pm 0.0055^*} $ \\
    ANIL~\cite{raghu2019rapid} & $ 0.7768\pm 0.0077^* $ & $ 0.7742\pm 0.0078^* $ & $ 0.7008\pm 0.0053^* $ & $ 0.7141\pm 0.0314^* $ & $ 0.7058\pm 0.0356^* $ & $ 0.6581\pm 0.0193^* $ \\
    BioBridge$_{Meta}$ & $ \bm{0.8621\pm 0.0023} $ & $ \bm{0.8663\pm 0.0019} $ & $ \bm{0.7693\pm 0.0028} $ & $ \bm{0.7919\pm 0.0072} $ & $ \bm{0.7910\pm 0.0084} $ & $ \bm{0.7130\pm 0.0056} $ \\
    \rowcolor{myco2} Setting                & \multicolumn{6}{c}{5-shot}                                                                                                                                                        \\
    MAML++~\cite{MAMLPP} & $ 0.8190\pm 0.0410 $ & $ 0.8173\pm 0.0450^* $ & $ 0.7390\pm 0.0305^* $ & $ 0.7190\pm 0.0382^* $ & $ 0.7077\pm 0.0461^* $ & $ 0.6614\pm 0.0273^* $ \\
    Protypes~\cite{snell2017prototypical} & $ 0.7895\pm 0.0020^* $ & $ 0.7886\pm 0.0003^* $ & $ 0.7105\pm 0.0038^* $ & $ 0.7393\pm 0.0239^* $ & $ 0.7330\pm 0.0259^* $ & $ 0.6730\pm 0.0173^* $ \\
    MetaOptNet~\cite{lee2019meta} & $ \uline{0.8561\pm 0.0142^*} $ & $ \uline{0.8388\pm 0.0162^*} $ & $ \uline{0.7834\pm 0.0128^*} $ & $ \uline{0.7748\pm 0.0010^*} $ & $ \uline{0.7537\pm 0.0059^*} $ & $ \uline{0.7143\pm 0.0049^*} $ \\
    ANIL~\cite{raghu2019rapid} & $ 0.8079\pm 0.0016^* $ & $ 0.8079\pm 0.0030^* $ & $ 0.7258\pm 0.0043^* $ & $ 0.7394\pm 0.0098^* $ & $ 0.7367\pm 0.0083^* $ & $ 0.6708\pm 0.0052^* $ \\
    BioBridge$_{Meta}$ & $ \bm{0.8802\pm 0.0025} $ & $ \bm{0.8828\pm 0.0030} $ & $ \bm{0.7894\pm 0.0026} $ & $ \bm{0.8249\pm 0.0149} $ & $ \bm{0.8254\pm 0.0145} $ & $ \bm{0.7446\pm 0.0126} $ \\
    \Xhline{1.2pt}      
    \rowcolor{white} \multicolumn{7}{l}{\small $*$ Significantly different (p < 0.05) from the corresponding BioBridge metric value; one-way analysis of variance (ANOVA).}\\                 
    \end{tabular}
    }
    \label{tab:a6}
\end{table*}

\subsection{Module ablation experiments}
The study presented in Table~\ref{tab:ablation_encode} evaluates the effectiveness of the Multi-scale Awareness, Stem, and GAU components within the BioBridge multi-level encoder. The removal of these elements demonstrated their individual contributions to the overall performance, with BioBridge consistently outperforming other models across various metrics. This underscores the importance of multi-level feature models in facilitating knowledge transfer, as they outshine their single-level counterparts.

\begin{figure}[!ht]
    \centering
    \includegraphics[width=\linewidth]{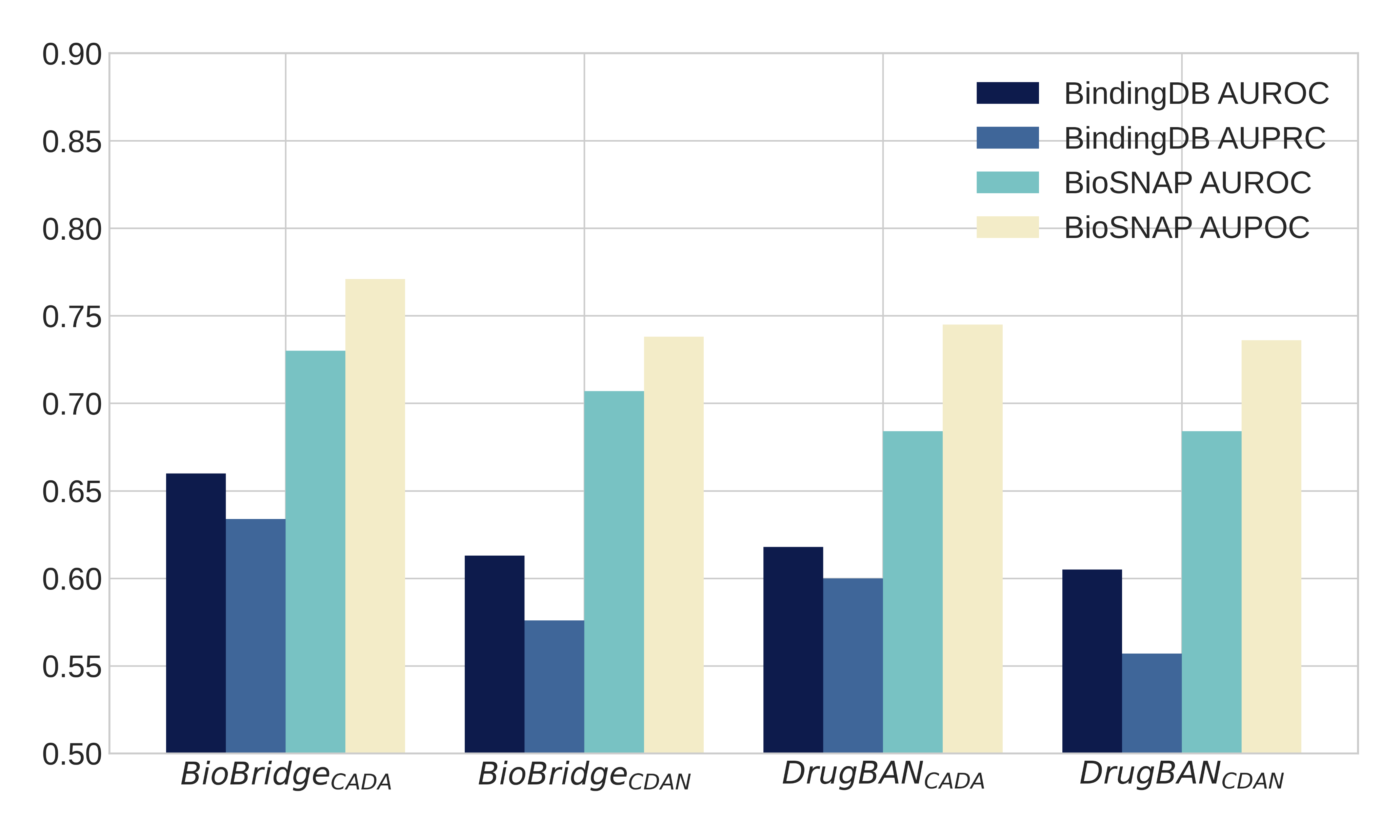}
    \caption{Ablation experiments for CADA modules in cross-domain settings.}
    \phantomsection
    \label{fig:cross_ablation}
\end{figure}

Fig~\ref{fig:cross_ablation} further examines the CADA module's performance in cross-domain tasks. It reveals that BioBridge when equipped with CADA, achieved the highest results. This is followed by DrugBAN with CADA and then BioBridge with CDAN. These findings indicate that a multi-level approach combined with understanding biological categories is crucial for accurately capturing drug-target interactions.

\begin{table*}[!ht]
    \renewcommand{\arraystretch}{1.2}
    \centering
    \rowcolors*{1}{myco}{white}
    \caption{Ablation experiments of BioBridge encoders on BindingDB and BioSNAP datasets (\textbf{Best},\uline{ Second Best}).}
    \phantomsection
    \label{tab:ablation_encoder}
    \resizebox{\linewidth}{!}{
    \begin{tabular}{ccccccccc}
    \Xhline{1.2pt}
    MA        & Stem      & GAU       & AUROC                       & AUPRC                       & ACC                         & AUROC                    & AUPRC                       & ACC                        \\ 
    \Xhline{1.2pt}
    \rowcolor{myco2}
    \multicolumn{3}{c}{Datasets}       & \multicolumn{3}{c}{BindingDB (Cold)}                                                    & \multicolumn{3}{c}{BioSNAP (Cold)}                                                  \\ 
    \ding{51} & \ding{51} & \ding{51} & \bm{$0.6799 \pm 0.0153$}    & \bm{$0.6271 \pm 0.0197$}    & \bm{$0.6339 \pm 0.0275$}    & \uline{$0.7160\pm 0.0109$}  & \bm{$0.7233\pm 0.0098$}     & \bm{$0.6648\pm 0.0187$}    \\
    \ding{55} & \ding{51} & \ding{51} & $0.6609 \pm 0.0121$         & $0.6077 \pm 0.0026$         & $0.5989 \pm 0.0180$          & $0.6896\pm 0.0068$       & $0.6912\pm 0.0102$          & $0.6262\pm 0.0102$         \\
    \ding{51} & \ding{55} & \ding{51} & $0.6336 \pm 0.0126$         & $0.5612 \pm 0.038$          & $0.5862 \pm 0.0352$         & $0.6863\pm 0.0510$       & $0.6927\pm 0.0683$          & $0.6436\pm 0.0314$         \\
    \ding{55} & \ding{55} & \ding{51} & \uline{$0.6767 \pm 0.0158$} & \uline{$0.6203 \pm 0.0061$} & $0.6267 \pm 0.0286$         & $0.6886\pm 0.0239$       & $0.6916\pm 0.0256$          & $0.6436\pm 0.0216$         \\
    \ding{51} & \ding{55} & \ding{55} & $0.6648 \pm 0.0023$         & $0.6022 \pm 0.0182$         & \uline{$0.6267 \pm 0.0083$} & \bm{$0.7189 \pm 0.0092$} & $0.7173 \pm 0.0126$         & $0.6637 \pm 0.0218$        \\
    \ding{55} & \ding{55} & \ding{55} & $0.6244 \pm 0.0215$         & $0.5802 \pm 0.0164$         & $0.5714 \pm 0.0452$         & $0.7112 \pm 0.0243$      & \uline{$0.7181 \pm 0.0192$} & \uline{$0.6639\pm 0.0165$} \\
    \multicolumn{3}{c}{$SADTI$}       & $0.6558 \pm 0.0122$         & $0.6002 \pm 0.0145$         & $0.6123 \pm 0.0232$         & $0.6619 \pm 0.0100$      & $0.6102 \pm 0.0145$         & $0.6223 \pm 0.0232$ \\
    \rowcolor{myco2}       
    \multicolumn{3}{c}{Datasets}       & \multicolumn{3}{c}{BindingDB (Random)}                                               & \multicolumn{3}{c}{BioSNAP (Random)}\\
    \ding{51} & \ding{51} & \ding{51} & \bm{$0.9660\pm 0.0009$}    & \bm{$0.9543\pm 0.0009$}    & \bm{$0.9143\pm 0.0032$}    & \bm{$0.9161\pm 0.0021$}    & \bm{$0.9192\pm 0.0031$}    & \bm{$0.8428\pm 0.0018$}    \\
    \ding{55} & \ding{51} & \ding{51} & \uline{$0.9637\pm 0.0022$} & $0.9512\pm 0.0030$         & $0.9091\pm 0.0007$         & $0.9124\pm 0.0021$         & $0.9152\pm 0.0035$         & \uline{$0.8420\pm 0.0006$} \\
    \ding{51} & \ding{55} & \ding{51} & $0.9636\pm 0.0014$         & \uline{$0.9516\pm 0.0014$} & \uline{$0.9123\pm 0.0016$} & \uline{$0.9127\pm 0.0020$} & \uline{$0.9174\pm 0.0029$} & $0.8381\pm 0.0048$         \\
    \ding{55} & \ding{55} & \ding{51} & $0.9625\pm 0.0013$         & $0.9500\pm 0.0024$         & $0.9031\pm 0.0013$         & $0.9105\pm 0.0030$         & $0.9121\pm 0.0039$         & $0.8368\pm 0.0050$         \\
    \ding{51} & \ding{55} & \ding{55} & $0.9632\pm 0.0006$         & $0.9498\pm 0.0039$         & $0.9086\pm 0.0015$         & $0.9093\pm 0.0019$         & $0.9118\pm 0.0041$         & $0.8391\pm 0.0046$         \\
    \ding{55} & \ding{55} & \ding{55} & $0.9618\pm 0.0030$         & $0.9508\pm 0.0035$         & $0.9050\pm 0.0090$         & $0.9090\pm 0.0044$         & $0.9105\pm 0.0053$         & $0.8374\pm 0.0068$         \\
    \multicolumn{3}{c}{$SADTI$}       & $0.9624\pm 0.0009$         & $0.9512\pm 0.0009$         & $0.9064\pm 0.0041$         & $0.9081\pm 0.0032$         & $0.9109\pm 0.0039$         & $0.8329\pm 0.0063$ \\
    \Xhline{1.2pt}
    \end{tabular}
    \label{tab:a7}
    }
\end{table*}
\subsection{BioBridge details}\label{sec:methods}
\begin{figure*}[h]
    \centering
    \includegraphics[width=0.9\textwidth]{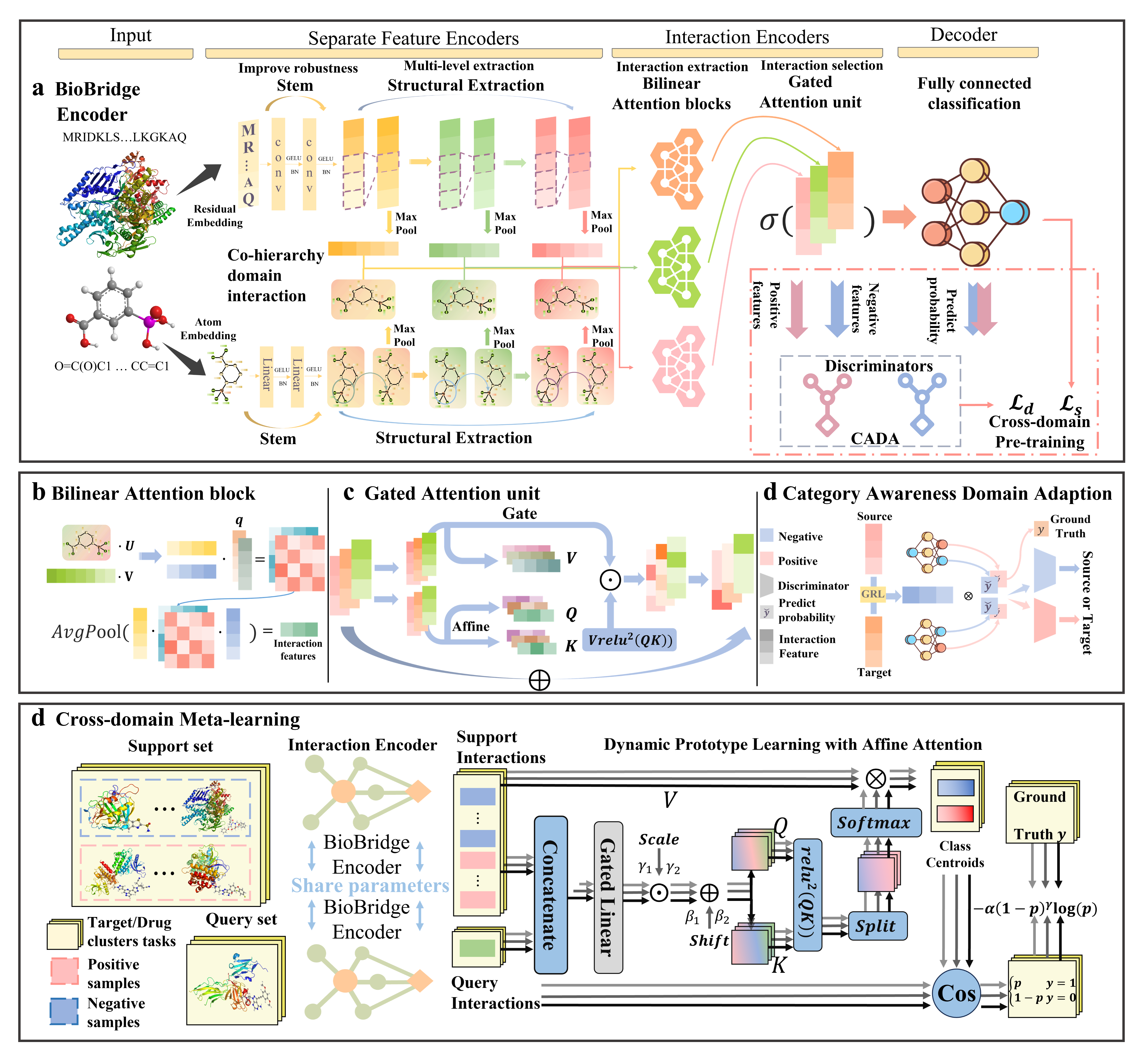}
    \caption{(a) BioBridge encoder processes protein sequences and molecular graphs using CNN and GCN to model their internal force. Bilinear attention captures drug-target interactions, generating diverse interaction fingerprints. These are integrated via gated attention units, with predictions made through a fully connected layer. In cross-domain pre-training, the CADA module enhances generalization through adversarial learning. (b) The bilinear attention network models drug-target interactions by multiplying drug and protein representations with transformation matrices $U$ and $V$, followed by low-rank bilinear interaction using $q$. The final interaction representation is obtained through average pooling. (c) The Gated Attention Unit processes input features through a learned gate and affine attention mechanism to produce enhanced output vectors. (d) CADA improves cross-domain generalization by embedding source and target domain representations and class probabilities into a joint representation. Separate discriminators minimize domain classification errors, enhancing domain distinction. (e) Meta-tasks are defined by clustering targets or drugs and dividing them into support and query sets. The BioBridge encoder generates interaction representations for these samples. An affine attention mechanism refines class prototypes by assessing the relative relationships within the support set. Cosine similarity then determines the query set's class. Focal Loss adaptively weights the learning difficulty of positive and negative samples.}
    \phantomsection
    \label{fig:8}
\end{figure*}

As depicted in Fig~\ref{fig:8}, the design details of BioBridge encompass a multi-level encoder along with its constituent modules and a dynamic prototype algorithm for cross-domain meta-learning.

The BioBridge encoder (Fig~\ref{fig:8}(a)) models the interaction forces between protein sequences and molecular graphs. Inspired by human-like multi-level understanding and the fact that features from shallow network layers are more readily transferable across domains compared to those from deeper layers~\cite{caron2021emerging,yosinski2014transferable}. Accordingly, BioBridge employs a multi-level aware encoder to capture a broader range of binding patterns. BioBridge enhances protein and molecule embeddings through a series of Stem layers, followed by a multi-layer extraction layer paired with a bilinear attention network to identify local drug-target interactions. A gated attention unit then integrates these interaction forces, enabling the model to predict the likelihood or affinity of protein-ligand interactions with a simple decoder.

\textbf{Input and Stem:} The input consists of target protein FASTA sequences and drug molecular graphs. These are processed through embedding layers to generate residue features $\mathcal{P}^{emb} \in \mathbb{R}^{\mathcal{L}_p \times \Theta_t}$ for proteins and atomic features $\mathcal{D}^{emb} \in \mathbb{R}^{\mathcal{L}_d \times \Theta_d}$ for drugs, where $\mathcal{L}_p$ and $\mathcal{L}_d$ represent the protein sequence length and the number of atoms, respectively, and $\Theta_t$ and $\Theta_d$ are the feature dimensions for proteins and atoms. To enhance structural information, we apply 1D convolutional layers to refine $\mathcal{P}^{emb}$, yielding $\mathcal{P}^{stem} \in \mathbb{R}^{\mathcal{L}_p \times \Theta_t}$, while drug graph features $\mathcal{D}^{emb}$ are processed through fully connected layers to produce $\mathcal{D}^{stem} \in \mathbb{R}^{\mathcal{L}_d \times \Theta_d}$.

\textbf{Structure Extractor:} The Protein Structural Extractor consists of three CNN blocks that generate multi-level features from the protein sequence. Each block treats the sequence as overlapping trimers (e.g., MRIDKS... GKAQ' $\to$ MRI', RID', IDK',...), using a $3 \times 3$ kernel, BatchNorm, MaxPool, and $1 \times 1$ convolutions to refine residue features.

The Drug Structural Extractor uses three GCN blocks to aggregate features from bonded atoms~\cite{kipf2016semi}, processing through GCN layers, max-pooling, and additional GCN layers for selecting local molecular interactions:
\begin{equation}
    \begin{aligned}
        &{\mathcal{D}}_{i}^{ex}=BN_i^{ex}(ReLu(GCN_i^{ex}({\mathcal{D}}_{i-1}^{ex}))) \\
        &{\mathcal{D}}_{i}^{out} = GCN_i^{out}(Maxpool_i^{out}({\mathcal{D}}_{i}^{ex})) \\
        &{\mathcal{P}}_{i}^{ex}=BN_i^{ex}(ReLu(CNN_i^{ex}({\mathcal{P}}_{i-1}^{ex}))) \\ 
        &{\mathcal{P}}_{i}^{out} = CNN_i^{out}(Maxpool_i^{out}({\mathcal{P}}_{i}^{ex}))\\
    \end{aligned}
\end{equation}

This is done for $i = 1, 2, 3$, with $\mathcal{D}_0^{ex} = \mathcal{D}^{stem}$ and $\mathcal{P}_0^{ex} = \mathcal{P}^{stem}$. Here, $\mathcal{P}_i^{ex}$ and $\mathcal{D}_i^{ex}$ are the extracted features for trimeric residues and drug atoms, respectively, while $\mathcal{P}_i^{out}$ and $\mathcal{D}_i^{out}$ represent the selected features at each level of the model.

\textbf{Bilinear Attention Network:}
To capture local interactions between drugs and proteins, we introduce bilinear attention modules within the same hierarchical model. These modules assign attention weights to each drug-protein interaction and pool these weights into feature vectors (Fig~\ref{fig:8}(b)).

At each level, protein and drug molecule features are represented as ${\mathcal{P}}_{i}^{out} ={p_i^1,p_i^2,\dots,p_i^N }$ and ${\mathcal{D}}_{i}^{out}={d_i^1,d_i^2,\dots,d_i^M }$, where $N$ and $M$ are the counts of selected protein structures and drug atoms, respectively. The interaction matrix $\mathbf{I}_i$ of size $N \times M$ is formulated as:
\begin{equation}
\begin{aligned}
\mathbf{I}_i = &((\mathbf{1}\cdot\mathbf{q_i}^{\mathsf{T}})\odot ReLu(({\mathcal{D}_{i}^{out}})^{\mathsf{T}}\mathbf{U}_i)) \cdot \\ 
&ReLu(\mathbf{V}_i^{\mathsf{T}}{\mathcal{P}_{i}^{out}}) + p_i
\end{aligned}
\end{equation}

Here, $V_i$ and $U_i$ are matrices that align protein and drug features into a shared space of dimension $dim$. We employ the Hadamard product and learn the weights $q\in \mathbb{R}^{k \times dim}$ and bias $p\in \mathbb{R}^{k \times 1}$ to generate $k$ interaction outcomes, emulating a multi-head attention mechanism.

The interaction outcomes are then converted into feature vectors for subsequent tasks, using the interaction matrix $I$ and variables $V_i, U_i$ to obtain $f_{i}^{\prime} \in \mathbb{R}^{k \times dim}$. Pooling operations are also used to simplify the features. Here $i\in \{ 1,2,3 \}$:

\begin{small}

\begin{equation}
     \begin{aligned}
 &f_{i}^{\prime} =ReLu(( {\mathcal{D}_{i}^{out}}^{\mathsf{T}}\mathbf{U}_i))^{\mathsf{T}} \cdot\mathbf{I}\cdot ReLu((\mathcal{P}_{i}^{out})^\mathsf{T}\mathbf{V_i}) \\
 &f_i=\mathrm{AvgPool}(\mathbf{f_i'},k)
 \end{aligned}
 \end{equation}

\end{small}

This method effectively captures multi-level drug-protein interactions, which helps to provide more diverse interaction features.

\textbf{Gated Attention Unit:}
We have engineered a streamlined gated unit with attention mechanisms to integrate interaction data across three hierarchical levels, as shown in Fig~\ref{fig:8}(c).

To integrate interaction data across three hierarchical levels, we design a gated attention unit, as shown in Fig~\ref{fig:8}(c). Interaction feature vectors $ f \in \mathbb{R}^{3 \times dim} $ are projected into a higher-dimensional space $Value \in \mathbb{R}^{3 \times d_h}$ using $ W_h \in \mathbb{R}^{dim \times d_h} $. A gate mechanism determines the significance of each feature via learned parameters, and the final feature vector is derived by adjusting the weighted interaction vectors with $ W_o \in \mathbb{R}^{d_h \times dim}$:

\begin{equation}
\begin{aligned}
&Value = SiLu(W_h f), gate = SiLu(W_g f), \\ 
& Z = SiLu(W_z f) \\
&Query = Z \gamma_1 + \beta_1, \quad Key = Z \gamma_2 + \beta_2
\end{aligned}
\end{equation}
Attention coefficients $ A \in \mathbb{R}^{3 \times 3} $ are calculated as $\frac{1}{3}\text{relu}^2(Query \cdot Key^{\mathsf{T}})$, and $ Value $ is multiplied by $ A $ to merge interaction vectors. The final feature vectors $ O \in \mathbb{R}^{dim} $ are selected by the $ gate $ through Hadamard multiplication with the enhanced interaction vectors and adjusted to the desired dimension using $ W_o \in \mathbb{R}^{d_h \times dim} $:
\begin{eqnarray}
A = \frac{1}{3}\text{relu}^2(Query \cdot Key^{\mathsf{T}}) \\
O = (Value \cdot A \odot gate) W_o
\end{eqnarray}
This approach enables nuanced representation across different levels, incorporating interaction patterns at all three model levels.

\end{appendices}

\end{document}